\documentclass{article}

\usepackage{PRIMEarxiv}

\usepackage[utf8]{inputenc} 
\usepackage[T1]{fontenc}    
\usepackage{hyperref}       
\usepackage{url}            
\usepackage{booktabs}       
\usepackage{amsfonts}
\usepackage{amsthm}%
\usepackage{float}
\usepackage{nicefrac}       
\usepackage{microtype}      
\usepackage{lipsum}
\usepackage{fancyhdr}       
\usepackage{graphicx}       
\graphicspath{{media/}}     
\usepackage{multirow}
\usepackage{multicol}
\usepackage{bm}
\usepackage{overpic}
\usepackage{cite}
\usepackage{subcaption}
\usepackage{amsmath}
\usepackage{amssymb}
\usepackage{comment}
\usepackage{algorithm}
\usepackage{algorithmicx}
\usepackage{algpseudocode}
\usepackage{color}
\usepackage{makecell}
\usepackage{comment}
\newcommand{\vect}[1]{\boldsymbol{#1}}

\title{Adversarial Fine-tuning in Offline-to-Online Reinforcement Learning for Robust Robot Control
}

\author{
  Shingo Ayabe \\
   Graduate School of Science and Engineering \\
   Chiba University \\
   Chiba, Japan \\
  \texttt{ayabe.shingo@chiba-u.jp} \\
   \And
  Hiroshi Kera\\
  Graduate School of Informatics \\
  Chiba University \\
  Chiba, Japan \\
  \texttt{kera@chiba-u.jp} \\
  \And
  Kazuhiko Kawamoto \\
  Graduate School of Informatics \\
  Chiba University \\
  Chiba, Japan \\
  \texttt{kawa@faculty.chiba-u.jp} \\
}

\begin{document}
\maketitle

\newtheorem{theorem}{Theorem}
\newtheorem{proposition}[theorem]{Proposition}%

\newtheorem{assumption}{Assumption}
\newtheorem{lemma}{Lemma}

\newtheorem{example}{Example}%
\newtheorem{remark}{Remark}%

\newtheorem{definition}{Definition}%

\begin{abstract}
Offline reinforcement learning enables sample-efficient policy acquisition without risky online interaction, yet policies trained on static datasets remain brittle under action-space perturbations such as actuator faults. This study introduces an offline-to-online framework that trains policies on clean data and then performs adversarial fine-tuning, where perturbations are injected into executed actions to induce compensatory behavior and improve resilience. A performance-aware curriculum further adjusts the perturbation probability during training via an exponential-moving-average signal, balancing robustness and stability throughout the learning process. Experiments on continuous-control locomotion tasks demonstrate that the proposed method consistently improves robustness over offline-only baselines and converges faster than training from scratch. Matching the fine-tuning and evaluation conditions yields the strongest robustness to action-space perturbations, while the adaptive curriculum strategy mitigates the degradation of nominal performance observed with the linear curriculum strategy. Overall, the results show that adversarial fine-tuning enables adaptive and robust control under uncertain environments, bridging the gap between offline efficiency and online adaptability.
\end{abstract}


\section{Introduction}
\label{sec:intro}
Offline reinforcement learning (offline RL)~\cite{offline_rl_1} trains policies from fixed datasets without direct interaction with the environment.
By avoiding online sampling, offline RL reduces data collection costs and potential safety risks. This benefit is particularly important in safety-critical domains such as healthcare~\cite{offline_rl_hc}, energy management~\cite{offline_rl_em}, and robot control~\cite{offline_rl_rc,offline_rl_2}.
However, standard RL algorithms often fail in offline settings due to overestimation and error accumulation in Q-values for out-of-distribution state-action pairs~\cite{BCQ}.
Conservative methods address this issue by constraining the learned policy to align with the behavior policy in the dataset~\cite{BCQ,IQL,TD3+BC,CQL}, which improves stability~\cite{offline_rl_2} but limits adaptability. 
Thus, robustness to unexpected disturbances not covered in fixed datasets remains a key requirement for real-world deployment.

\begin{figure}[t]
    \centering
     \includegraphics[width=1.0\hsize]{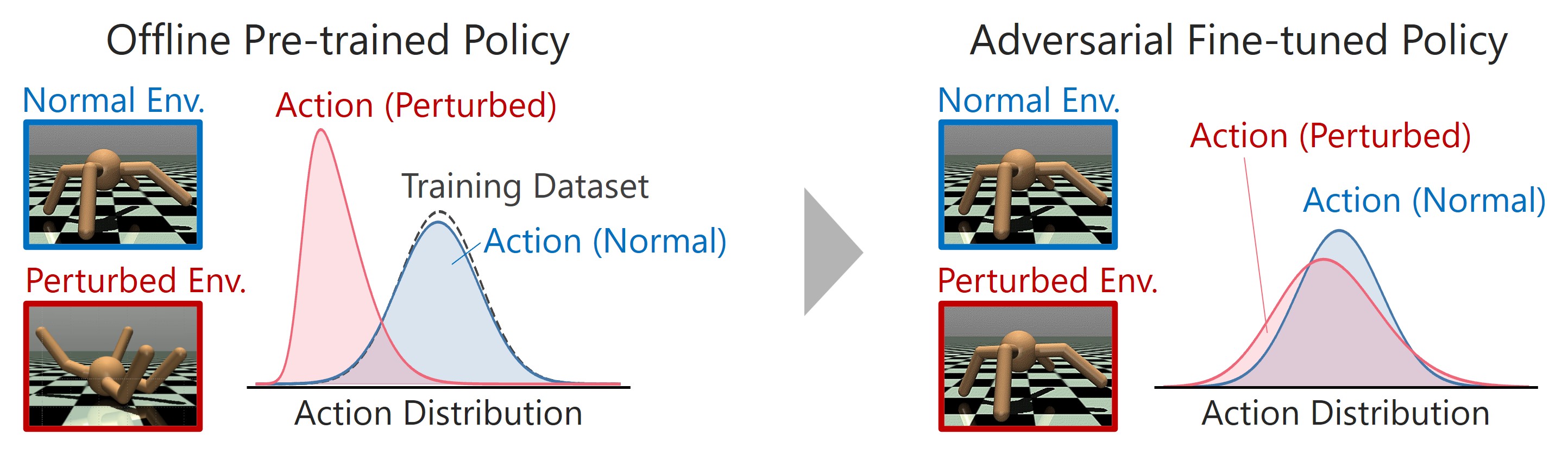}
     \caption{Incompatibility between conservative offline RL and action-space robustness. Left: A conservatively trained policy matches the dataset (dashed line) under normal conditions (blue) but fails under perturbations (red). Right: Robustness requires compensatory actions that maintain consistent distributions across both conditions (blue and red overlap).}
    \label{fig:concept_overview}
\end{figure}

Our key insight is that conservative offline RL and action-space robustness are incompatible. 
Conservative methods constrain policies to dataset actions to prevent extrapolation errors, yet robustness to action perturbations requires learning from the out-of-distribution samples that these constraints prohibit.
This incompatibility is illustrated in Figure~\ref{fig:concept_overview}. 
In the left panel, a conservatively trained policy produces an action distribution (blue) that aligns with the offline dataset (dashed) under normal conditions.
However, when action perturbations 
occur, the executed actions shift into out-of-distribution regions (red) 
where the policy fails. 
The right panel illustrates that if the policy could learn to produce 
compensatory actions that keep the executed action distribution consistent 
across both normal and perturbed conditions, robustness would be achieved 
(blue and red overlap).

Building on this insight, we formalize adversarial fine-tuning within an offline-to-online
framework~\cite{AWAC,BR,adaptive_bc,PEX}. 
We first pretrain a policy on clean data for sample efficiency. 
Then, during online fine-tuning, we deliberately inject adversarial perturbations, enabling the policy to learn compensatory behaviors for perturbed conditions. 
Furthermore, we introduce a curriculum learning mechanism that gradually increases the perturbation probability based on training progress or policy performance.
This curriculum balances learning difficulty and stability, preventing catastrophic performance drops while guiding the policy toward robustness in both normal and perturbed scenarios.

However, existing work on robustness in offline RL has focused primarily on
distributional shifts in the state-space, leaving this incompatibility largely unaddressed. 
Most prior work addresses state perturbations (e.g., sensor noise) through smoothness 
penalties and output stabilization~\cite{RORL,MICRO,robust_testtime_5}, 
which remain compatible with conservative constraints. In contrast, action 
perturbations (e.g., actuator faults) have been studied far less. These 
disturbances directly modify executed actions, and recent work shows that 
offline RL is particularly vulnerable to such perturbations~\cite{offline_robust}, as datasets typically lack perturbation-aware transitions 
while conservative constraints prevent necessary exploration.

To validate our approach, we conduct experiments in legged robot environments from OpenAI Gym~\cite{OpenAI},
where either random or adversarial perturbations are added to joint torque commands to simulate actuator faults.
The proposed method is compared with two baselines: offline-only training and fully online training, and it consistently achieves superior robustness across all environments.
In addition, a comparison of linear and adaptive schedules shows that the adaptive approach maintains robustness while preserving training stability.

The contributions of this work are summarized as follows:
\begin{itemize}
\item We propose an adversarial fine-tuning method that injects perturbations during online training. This enables targeted adaptation to action perturbations while preserving the sample efficiency of offline pretraining.
\item We demonstrate that adversarial fine-tuning consistently outperforms offline-only and fully online baselines in terms of robustness.
\item We propose an adaptive curriculum that adjusts perturbation probability based on policy performance. This prevents overfitting to adversarial conditions while maintaining training stability, addressing a key limitation of fixed-schedule approaches.
\end{itemize}

\section{Related Work}
\subsection{Offline RL}
Offline RL is a framework that trains policies using pre-collected datasets without interaction with the environment.
In the lack of exploration, directly applying conventional online RL algorithms in this setting often leads to overestimation of Q-values for out-of-distribution state-action pairs, which may cause unstable learning~\cite{offline_rl_2}.
To address this issue, conservative approaches have been developed to keep the learned policy close to the data distribution~\cite{BCQ,TD3+BC,CQL,IQL,Fisher-BRC}.
For instance, TD3+BC~\cite{TD3+BC} enhances offline training efficiency by incorporating a simple behavior cloning loss into the actor update of TD3~\cite{TD3}.
These algorithms provide a foundation for offline policy training but remain limited by their reliance on fixed datasets.

\subsection{Robust RL}
Robustness under uncertainty has been widely studied not only in RL but also in other domains such as cybersecurity and remote sensing~\cite{robust_cross_1, robust_cross_2}, where models must operate reliably under environmental variability or adversarial conditions. 
Within RL, robustness has also been explored through ensemble-based approaches~\cite{emsemble_1, emsemble_2, emsemble_3}, which combine multiple models or representations to reduce sensitivity to input variability and distributional shifts. 
These approaches mainly aim to improve stability and generalization during training under uncertain inputs.
Ensemble-based or representation-level approaches primarily address robustness to input variability during training.
In contrast, the present study focuses on robustness against action-space perturbations that occur at execution time in control tasks.

In RL for robot control, robustness is crucial for handling uncertainties such as sensor noise, matched physical parameter, unexpected terrain variations, or mechanical failures~\cite{state_perturbation_1,state_perturbation_2,adv_trng_1,o2022_adv,action_perturbation_1}.
A common approach is to introduce perturbations during training to improve policy generalization across conditions.
Domain randomization~\cite{domain_random_1,domain_random_2,domain_random_3} improves adaptability by randomizing environment parameters such as friction, while adversarial training~\cite{adv_trng_1,adv_trng_2,adv_trng_3,adv_trng_4,adv_trng_5} improves robustness by exposing policies to adversarial perturbations.

In offline RL, robustness has also been studied, mainly against state-space perturbations.
RORL~\cite{RORL} improves robustness by smoothing the value distribution and applying conservative estimation, and Nguyen et al.~\cite{robust_testtime_5} propose a defense mechanism based on adversarial perturbations with KL divergence.
These methods demonstrate progress in improving robustness against state perturbations.
In contrast, robustness against action-space perturbations remains less explored.
Recent studies have applied an adversarial action-space perturbation generation method based on differential evolution~\cite{o2022_adv} to offline-trained policies~\cite{offline_robust}.
The results demonstrate that offline-trained policies are more vulnerable to such perturbations than online-trained policies.
However, effective methods to improve robustness against such perturbations have not yet been established.
The difficulty arises because conservative constraints in offline RL suppress out-of-distribution actions.
As a result, adaptation is impossible unless perturbation-adapted actions and outcomes are already present in the dataset.
Moreover, the lack of online interaction makes it impossible to acquire such data during training.
These characteristics make robustness to action-space perturbations a particularly challenging problem, motivating the present study.

\subsection{Offline-to-Online RL}
Offline-to-online RL has emerged as a promising framework to overcome the limitations of offline RL, which is highly dependent on dataset quality and lacks adaptability to unseen disturbances~\cite{AWAC,BR,adaptive_bc,PEX,OEMA,FamO2O}.
This framework first pretrains a policy on fixed datasets and then fine-tunes it online through direct interaction with the environment.
By combining the efficiency of offline pretraining with the adaptability of online training, offline-to-online RL provides a way to address action-space perturbations that cannot be handled in purely offline settings.

A major challenge in offline-to-online RL is performance degradation in the early phase of fine-tuning, caused by distributional shifts between offline data and newly collected online experiences~\cite{AWAC,BR}.
Various approaches have been proposed to mitigate this issue.
Representative examples include adaptive or relaxed policy constraints~\cite{AWAC,APL,adaptive_bc}, data selection methods that prioritize near-on-policy samples~\cite{BR}, dynamic switching among multiple policies depending on the state~\cite{PEX,FamO2O}, and exploration strategies combined with meta-adaptation~\cite{OEMA}.
These offline-to-online RL methods primarily consider distributional shifts under nominal environments, where the mismatch arises from differences between offline data and newly collected online experiences during data collection.
In contrast, this study focuses on action-space perturbations, representing discrepancies between training and deployment environments.

This study leverages offline-to-online RL to improve robustness against action-space perturbations.
To address this challenge, the policy is first pretrained offline and then fine-tuned in an online environment where perturbations are introduced.
This approach enables policies to acquire robustness to action-space perturbations adaptively, which is difficult to achieve in purely offline settings.

\section{Preliminaries}
This section provides background on reinforcement learning methods used in this study. 
We first introduce the formulation of offline reinforcement learning (RL) and discuss its limitations under action-space perturbations. 
We then describe the offline-to-online RL framework to address these limitations.

\subsection{Offline RL}
\label{sec:3.1}
Reinforcement learning (RL) is formulated as a Markov decision process (MDP), represented by the tuple $(\mathcal{S}, \mathcal{A}, P, P_0, r, \gamma)$. Here, $\mathcal{S}$ denotes the state space, $\mathcal{A}$ the action space, $P(\vect{s}'|\vect{s}, \vect{a})$ the state transition probability, $P_0$ the initial state distribution, $r(\vect{s}, \vect{a})$ the reward function, and $\gamma$ the discount factor. The objective is to find an optimal policy $\pi^*$ that maximizes the expected return.

Offline RL trains policies using a fixed dataset $\mathcal{D}=\{(\vect{s}_t, \vect{a}_t, \vect{s}_{t+1}, r_t)_i\}$ without further interaction with the environment. The dataset is usually collected from expert demonstrations, past policy rollouts, or random actions, and is assumed to follow a behavior policy $\pi_\beta$. A common difficulty in this setting is overestimation of Q-values for out-of-distribution state-action pairs. Without exploration, these errors remain uncorrected and lead to poor generalization. Conservative learning mitigates this problem by constraining the learned policy to remain close to the behavior policy. The objective can be written as:
\begin{equation}
    \label{equ:offline_rl}
    \pi = \arg\max_{\pi} \mathbb{E}_{\vect{s}_t \sim \mathcal{D}} \left[ Q^\pi(\vect{s}_t, \pi(\vect{s}_t)) \right] \quad \text{s.t. } D(\pi(\cdot|\vect{s}_t), \pi_\beta(\cdot|\vect{s}_t)) < \epsilon,
\end{equation}
where $D$ denotes a divergence metric such as the KL divergence~\cite{offline_rl_2}.

Such conservative designs, however, create vulnerabilities when perturbations affect the executed actions. In real systems, the executed action may deviate from the intended policy output due to disturbances or noise, which we denote as the perturbed action $\tilde{\vect{a}}_t$.
Consequently, both the reward $\tilde{r}_t=r(\vect{s}_t,\tilde{\vect{a}}_t)$ and the next state $\tilde{\vect{s}}_{t+1}\sim P(\cdot \mid \vect{s}_t, \tilde{\vect{a}}_t)$ depend on $\tilde{\vect{a}}_t$.

Offline datasets typically contain unperturbed actions $\vect{a}$, making it impossible to assess whether policies remain effective once perturbations are applied. Collecting transitions with perturbation-adapted actions and outcomes is practically infeasible without online interaction. In addition, the conservative nature of offline RL may inadvertently reinforce actions that are fragile under perturbations. These limitations expose a structural vulnerability of offline RL to action-space perturbations, stemming from both the lack of robust data and the restrictive training paradigm.
Therefore, new frameworks that can actively adapt to perturbations are required.

\subsection{Offline-to-Online RL}
\label{sec:3.2}
Offline-to-online RL~\cite{AWAC,BR} has emerged as a practical framework to address the limitations of offline RL. 
The process consists of two stages:  
(1) offline pretraining, where a policy is initialized using fixed datasets, and  
(2) online fine-tuning, where the policy continues to learn through interactions with the target environment.  

In this study, offline pretraining is performed with TD3+BC~\cite{TD3+BC}, a conservative actor–critic algorithm. 
The policy update rule is given by
\begin{equation}
    \pi = \arg\max_{\pi} \mathbb{E}_{(\vect{s}_t,\vect{a}_t)\sim \mathcal{D}}
    \Bigl[ Q^\pi(\vect{s}_t,\pi(\vect{s}_t)) - \|\pi(\vect{s}_t)-\vect{a}_t\|^2 \Bigr],
\end{equation}
where the second term enforces conservatism by penalizing deviation from the behavior policy.

During online fine-tuning, the constraint is removed and the policy is updated using the standard TD3 rule~\cite{TD3}:
\begin{equation}
    \label{equ:td3_update}
    \pi = \arg\max_{\pi} \mathbb{E}_{\vect{s}_t \sim \mathcal{R}} \left[ Q^\pi(\vect{s}_t, \pi(\vect{s}_t)) \right],
\end{equation}
where $\mathcal{R}$ denotes the replay buffer containing online experience.
Given a transition $(\vect{s}_t,\vect{a}_t,\vect{s}_{t+1},r_t)$ from 
$\mathcal{R}$, the target for the critics is
\begin{equation}
    \label{equ:critic_target}
    y_t = r_t + \gamma \min_{i\in\{1,2\}} Q_{\theta_i^-}\!\Bigl(\vect{s}_{t+1},
    \pi_{\phi^-}(\vect{s}_{t+1})+\varepsilon\Bigr),
\end{equation}
where $\gamma$ is the discount factor, $Q_{\theta_i^-}$ and $\pi_{\phi^-}$ denote the target critic and target policy, respectively, and $\varepsilon\sim\mathrm{clip}(\mathcal{N}(0,\sigma^2),-c,c)$ is the target policy smoothing noise. The operator $\mathrm{clip}(x, a, b)$ restricts the value $x$ to the interval $[a, b]$.
The critics minimize the Bellman error:
\begin{equation}
    \label{equ:critic_loss}
    \mathcal{L}(\theta_i) = \mathbb{E}\bigl[(Q_{\theta_i}(\vect{s}_t,\vect{a}_t)-y_t)^2\bigr],\quad i=1,2.
\end{equation}

Some prior work~\cite{APL,adaptive_bc} retains the TD3+BC penalty during fine-tuning. 
However, when perturbations are introduced, the effectiveness of stored actions under perturbed dynamics is not guaranteed. 
In such cases, maintaining the constraint may hinder adaptation. 
Therefore, we discard the penalty during online fine-tuning to allow more flexible policy updates.
The Appendix~\ref{appdx_hyperparams} presents a comparison of final performance with and without the BC penalty.

\subsection{Limitation of Conservative Offline RL under Action-space Perturbations}
\label{sec:3.3}
This section clarifies under what condition the conservative constraint in Eq.~\ref{equ:offline_rl} can limit compensation for action-space perturbations introduced in Section~\ref{sec:3.1}.

\subsection*{Sufficient Condition}
The following condition describes when the conservative constraint can prevent compensation.

\begin{lemma}[Sufficient condition for constraint-induced limitation]
    Suppose that maintaining performance under the perturbed environment requires intended actions that lie outside an $\epsilon$-neighborhood of $\pi_\beta(\cdot|\vect{s})$ for some states $\vect{s}$.
    Then any offline RL method enforcing $D(\pi(\cdot|\vect{s}),\pi_\beta(\cdot|\vect{s})) \leq \epsilon$ cannot generally be expected to learn such compensation from the dataset $\mathcal{D}$ alone.
\end{lemma}

\begin{proof}[Proof sketch]
    The dataset $\mathcal{D}$ contains only clean transitions.
    It provides no supervision signal for perturbed outcomes.
    Meanwhile, the divergence constraint restricts policy updates within the $\epsilon$-neighborhood.
    If compensation requires a larger deviation, the update is explicitly restricted.
    Therefore, compensation cannot typically be learned under this constraint without additional online interaction.
\end{proof}

\subsection*{Interpretation and Practical Implication}

This condition does not imply that robustness is impossible in all offline settings.
If the offline dataset already covers the necessary compensatory actions, robustness may still be achievable offline.
However, when perturbations are adversarial, they are designed to reduce performance under the nominal policy.
Compensation then tends to require policy adjustments toward directions where the nominal policy performs poorly.
Such directions often lie outside the high-density region of the behavior distribution.
Even if they remain within the support, they may require shifting probability mass toward low-density regions.

such probability shifts are discouraged by conservative objectives.
As a result, these conservative constraints are likely to become effectively binding along the directions required for compensation.
This mechanism helps explain why additional online interaction can be necessary.
Online data provide transitions under perturbed actions and can reduce the effective limitation imposed by purely conservative offline objectives.

\section{Proposed Method}
\label{4}
This study addresses the problem of improving policy robustness against action-space perturbations in the offline RL setting. 
As discussed in Section~\ref{sec:3.1}, offline RL constrains policies to stay close to actions in the dataset, as shown in Equation~(\ref{equ:offline_rl}). 
Thus, when the dataset lacks transitions with perturbed actions, it is difficult to obtain a policy that adapts to perturbed environments. 
To solve this limitation, we adopt an offline-to-online approach, pretraining a policy on a clean offline dataset and fine-tuning it in online adversarial environments.

\subsection{Problem Formulation}
\label{4.1}
We formulate the problem of robust policy training under action-space perturbations. 
The objective is to find an optimal policy $\pi^*$ that maximizes the expected cumulative reward under adversarial perturbations:
\begin{equation}
    \pi^* = \arg\max_{\pi}\;\min_{\tilde{\vect{a}} \in \mathcal{U}}
    \mathbb{E}\!\left[\sum_{t=0}^\infty \gamma^t r(\vect{s}_t, \tilde{\vect{a}})\right],
    \label{equ: adversarial training}
\end{equation}
where $\tilde{\vect{a}}$ is an adversarially perturbed action chosen from a predefined set $\mathcal{U}$.

In our approach, the initial policy for this optimization is provided by offline training on a clean dataset, which improves training efficiency during online fine-tuning. 
We then approximate the intractable min–max objective within the TD3-based actor–critic framework, described in Section~\ref{sec:3.1}.
In this framework, the collected transitions~$(\vect{s}_t,\vect{a}_t,\tilde{r}_t,\tilde{\vect{s}}_{t+1})$ are stored in the replay buffer and used to compute the critic target:
\begin{equation}
    \label{equ:critic_target_2}
    y_t = \tilde{r}_t + \gamma \min_{i\in\{1,2\}}
    Q_{\theta_i^-}\!\Bigl(\tilde{\vect{s}}_{t+1},
    \pi_{\phi^-}(\tilde{\vect{s}}_{t+1})+\varepsilon\Bigr),
\end{equation}
with $\tilde{\vect{s}}_{t+1} \sim P(\cdot \mid \vect{s}_t, \tilde{\vect{a}}_t)$ and $\tilde{r}_t = r(\vect{s}_t, \tilde{\vect{a}}_t)$.

Through repeated actor–critic updates, the policy gradually acquires robustness to perturbations during online fine-tuning.
Building on this setting, the next section introduces our adversarial fine-tuning method.

\begin{figure}[t]
    \centering
     \includegraphics[width=1.0\hsize]{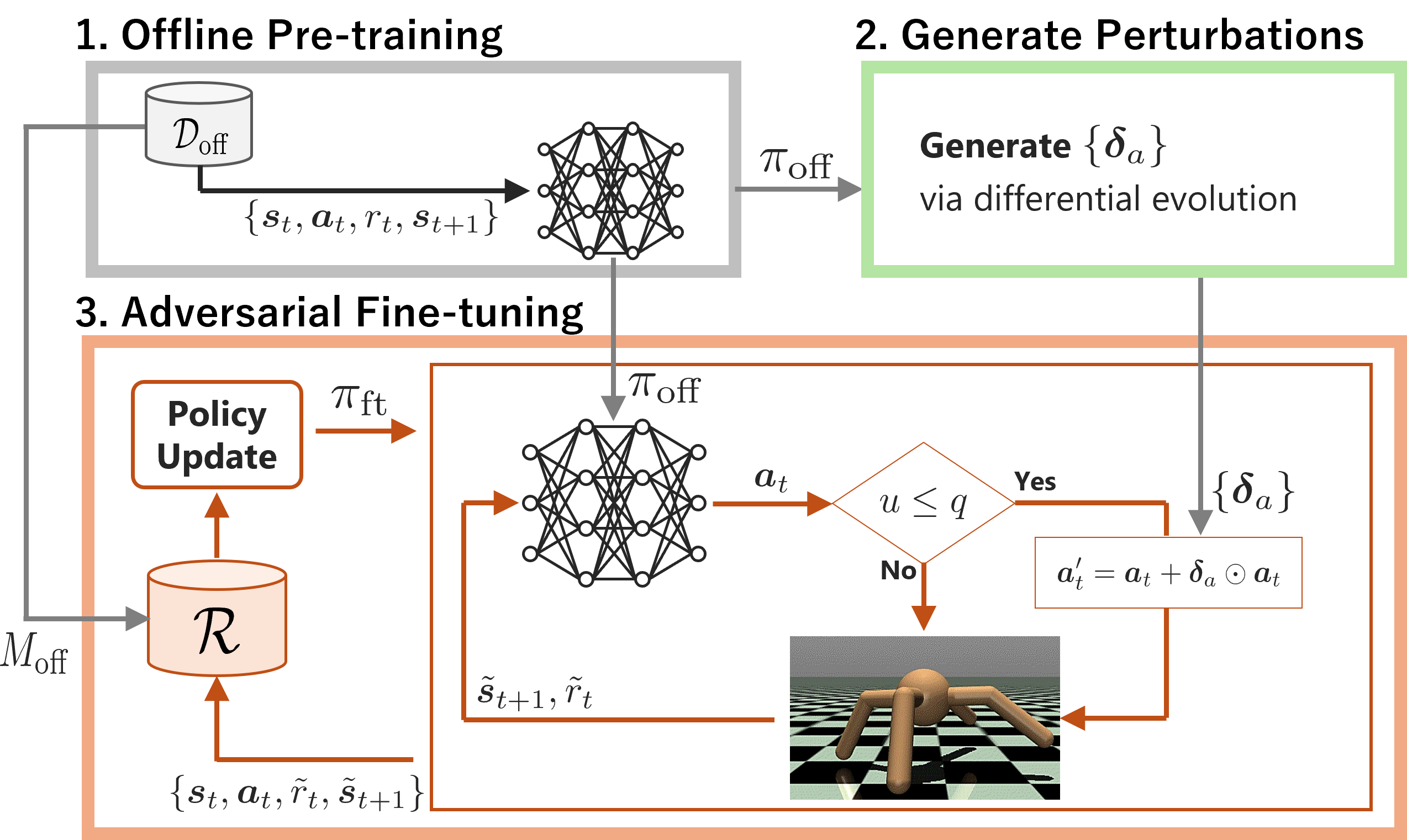}
    \caption{Overview of the proposed framework with three stages:
    (1) {Offline Pretraining}, where a policy $\pi_\text{off}$ is trained on a clean dataset;
    (2) {Adversarial Perturbation Generation}, where perturbations $\{\delta_a\}$ are precomputed with $\pi_\text{off}$ before fine-tuning;
    and (3) {Adversarial Fine-Tuning}, where $\pi_\text{off}$ is refined in perturbed environments using these perturbations and replay buffer updates.}
    \label{fig:overview}
\end{figure}

\subsection{Adversarial Fine-tuning}
\label{4.2}
Adversarial fine-tuning adapts offline-pretrained policies to perturbed environments by injecting adversarial perturbations with a probability $q$ per episode. 
An overview of the overall framework is illustrated in Figure~\ref{fig:overview}. 

The policy $\pi_\text{ft}$ and critic $Q_\text{ft}$ are initialized from offline-pretrained models $\pi_\text{off}$ and $Q_\text{off}$. 
To stabilize fine-tuning, a portion of the offline dataset $\mathcal{D}_\text{off}$ is inserted into the replay buffer $\mathcal{R}$ at the beginning of training, following prior works~\cite{BR,APL,adaptive_bc}. 
A predefined ratio of offline samples is used for initialization so that training starts from a balanced mixture of offline and online experiences, reducing instability during early fine-tuning.

In each episode, the agent samples actions from the current policy and perturbs them with probability $q$:
\begin{equation}
\label{equ:attack}
    \tilde{\vect{a}}_t = \vect{a}_t + \vect{\delta}_a \odot \vect{a}_t
    \quad \text{with probability } q,
\end{equation}
where $\odot$ denotes the element-wise (Hadamard) product and $\vect{\delta}_a$ is an adversarial perturbation. 
Before fine-tuning, a set of perturbations $\{\vect{\delta}_a\}$ is pre-generated using the differential evolution (DE) algorithm~\cite{o2022_adv,offline_robust} with respect to the offline-pretrained policy. 
This precomputation avoids the costly inner-loop minimization required by conventional adversarial training during fine-tuning.

Algorithms~\ref{algo:ft_fixed} and \ref{algo:episode_loop}
summarize the overall procedure for the fixed-$q$ case.
In Section \ref{4.3}, we extend this algorithm to introduce curriculum schedules that adaptively update $q$ over time.
\begin{algorithm}[t]
\caption{Adversarial Fine-Tuning}
\label{algo:ft_fixed}
\begin{algorithmic}[1]
    \Require $\pi_\text{off}, Q_\text{off}$: offline-pretrained policy and critic
    \Require $\mathcal{D}_\text{off}$: offline dataset
    \Require $q\in[0,1]$: fixed perturbation probability
    \State Initialize $\pi_\text{ft}, Q_\text{ft} \gets \pi_\text{off}, Q_\text{off}$
    \State Initialize replay buffer $\mathcal{R}$ with a fraction of offline samples
    \While{within training budget}
        \State Generate perturbation $\vect{\delta}$ using Eq.~(\ref{equ:perturbations_case})
        \State $\pi_\text{ft} \gets \textsc{EpisodeLoop}(\pi_\text{ft}, Q_\text{ft}, \vect{\delta}, q)$
    \EndWhile
\end{algorithmic}
\end{algorithm}
\begin{algorithm}[t]
\caption{\textsc{EpisodeLoop} $(\pi_\text{ft}, Q_\text{ft}, \vect{\delta}, q)$}
\label{algo:episode_loop}
\begin{algorithmic}[1]
    \State Sample initial state $\vect{s}_0 \sim P_0(\cdot)$
    \While{episode not terminated}
        \State Sample action $\vect{a} \sim \pi_\text{ft}(\cdot \mid \vect{s})$
        \State $\vect{a} \gets \vect{a} + \vect{\delta} \odot \vect{a}$ with probability $q$
        \State Optimize $\pi_\text{ft}$ and $Q_\text{ft}$
    \EndWhile
    \State \Return $\pi_\text{ft}$
\end{algorithmic}
\end{algorithm}

\subsection{Curriculum-based Fine-tuning}
\label{4.3}
While fixed-$q$ fine-tuning (Section~\ref{4.2}) provides a stable baseline, it cannot account for changes in training dynamics or difficulty across episodes. 
We therefore introduce two curriculum-based extensions that gradually adjust the perturbation probability $q$: a \emph{linear curriculum} and an \emph{adaptive curriculum}. 
Both use the same general update rule:
\begin{equation}
    \label{equ:q_update}
    q \;\gets\; \mathrm{clip}\!\left(q + \Delta q,\, 0,\, 1\right),
\end{equation}
where $\Delta q$ represents the increment per update, and \(\mathrm{clip}(\cdot)\) ensures that $q$ remains within $[0,1]$.

\subsubsection{Linear curriculum}
The linear schedule increases $q$ at a constant rate throughout fine-tuning.
Starting from $q_\text{init}$, the perturbation probability is linearly raised toward $q_\text{max}$ as training progresses.
This gradual increase encourages the agent to first adapt to relatively mild perturbations and later to stronger ones, improving robustness without destabilizing early learning.
The update rule simply uses a fixed increment, 
\begin{equation}
    \label{equ:linear_curriculum}
    \Delta q = c,    
\end{equation}
where $c$ is a constant step size.
Although straightforward, overly large $c$ may lead to excessive exposure to perturbations too early, which can degrade nominal performance under normal (unperturbed) conditions (see Appendix~\ref{appdx_schedule}).

\subsubsection{Adaptive curriculum}
\label{sec:adaptive_curriculum}
The adaptive curriculum dynamically adjusts the perturbation probability $q$ according to the policy's performance.
When the policy shows improvement, the curriculum increases \(q\) to enhance robustness.
When the performance declines, the curriculum decreases \(q\)
to reduce task difficulty and stabilize training.
Through this adaptive mechanism, the agent maintains a balance
between robustness and stability during fine-tuning.

The overall fine-tuning process using the adaptive curriculum is summarized in Algorithm~\ref{algo:algo_ac}.
This algorithm extends Algorithm~\ref{algo:ft_fixed} by introducing performance-based updates of the perturbation probability $q$ through periodic evaluation.
During training, the current policy is periodically evaluated across multiple episodes, and the average normalized score $R_n$ is used as a measure of policy performance.
The evaluation process is described in detail in Algorithm~\ref{algo:policy_eval}.
\begin{algorithm}[t]
\caption{Adversarial Fine-Tuning with Adaptive Curriculum}
\label{algo:algo_ac}
\begin{algorithmic}[1]
    \Require $\pi_\text{off}, Q_\text{off}$: offline-pretrained policy and critic
    \Require $\mathcal{D}_\text{off}$: offline dataset
    \Require $q_\text{init}\in[0,1]$: initial perturbation probability
    \State Initialize $\pi_\text{ft}, Q_\text{ft} \gets \pi_\text{off}, Q_\text{off}$
    \State Initialize replay buffer $\mathcal{R}$ with a fraction of offline samples
    \State $q \gets q_\text{init},n\gets 0$
    \While{within training budget}
        \State Generate perturbation $\vect{\delta}$ using Eq.~(\ref{equ:perturbations_case})
        \State $\pi_\text{ft} \gets \textsc{EpisodeLoop}(\pi_\text{ft}, Q_\text{ft}, \vect{\delta}, q)$
        \If{at regular evaluation intervals}
            \State $R_n \gets \textsc{PolicyEvaluation}(\pi_\text{ft}, q,n)$
            \State Update $q$ using Eqs.~(\ref{equ:q_update}) and~(\ref{equ:adaptive_schedule})
            \State $n\gets n+1$
        \EndIf
    \EndWhile
\end{algorithmic}
\end{algorithm}

The update rule modifies the increment \(\Delta q\)
based on the change in the exponentially smoothed evaluation score:
\begin{equation}
\label{equ:adaptive_schedule}
\Delta q = \eta \, (\bar{R}_n - \bar{R}_{n-1}),
\quad
\text{where }
\bar{R}_n = \beta R_n + (1 - \beta)\bar{R}_{n-1}.
\end{equation}
The exponential moving average  \(\bar{R}_n\) provides a stable estimate of performance trends by filtering out short-term fluctuations.
The parameters \(\eta\) and \(\beta\) control the adaptation dynamics: a smaller \(\beta\) captures long-term stability, while a larger \(\beta\) responds more quickly to recent changes.
A larger \(\eta\) leads to stronger adaptation, whereas a smaller \(\eta\) results in smoother updates.
By adjusting these parameters, the adaptive curriculum gradually modifies \(q\) to improve robustness
without causing instability.
Theoretical analysis is provided in Appendix~\ref{appdx_theory}, and the settings of $\eta$ and $\beta$ are detailed in Appendix~\ref{appdx_schedule}.

\begin{algorithm}[t]
\caption{\textsc{PolicyEvaluation} ($\pi_\text{ft}, q,n$)}
\label{algo:policy_eval}
\begin{algorithmic}[1]
    \Require $M$: total number of evaluation episodes
    \For{each episode $m = 1,2,\ldots,M$}
        \State Generate perturbation $\vect{\delta}$ using Eq.~(\ref{equ:perturbations_case})
        \State Sample initial state $\vect{s}_0 \sim P_0(\cdot)$
        \State $R^{(m)} \gets 0$
        \While{episode not terminated}
            \State Sample action $\vect{a} \sim \pi_\text{ft}(\cdot \mid \vect{s})$
            \State $\vect{a} \gets \vect{a} + \vect{\delta} \odot \vect{a}$ with probability $q$
            \State Execute $\vect{a}$, observe reward $r$, and update $R^{(m)} \gets R^{(m)} + r$
        \EndWhile
    \EndFor
    \State Compute $R_n \gets \frac{1}{M}\sum^{M}_{m=1} R^{(m)}$ 
    \State \Return $R_n$
\end{algorithmic}
\end{algorithm}

\section{Experiments}
\label{5}
This section presents experimental evaluations of the proposed adversarial fine-tuning framework. 
First, we describe the experimental setup, including pretrained models, fine-tuning conditions, perturbation configurations, and evaluation protocols. 
Next, we compare fine-tuned policies with offline and fully online baselines to evaluate robustness under different perturbation conditions. 
Finally, we examine curriculum fine-tuning strategies that vary the perturbation probability $q$ during training and analyze their impact on robustness against adversarial action-space perturbations.

\subsection{Experimental Setup}
\label{5.1}
\noindent
\textbf{Pretrained Model}
The policy is pretrained for 5 million steps using TD3+BC~\cite{TD3+BC}, an offline reinforcement learning algorithm, implemented with the d3rlpy library~\cite{d3rlpy}. The expert dataset from Datasets for Deep Data-Driven Reinforcement Learning (D4RL)~\cite{D4RL} is used for training.

\noindent
\textbf{Fine-Tuning Settings}
Fine-tuning is conducted for 1 million steps using the TD3 algorithm. Each configuration is trained with 5 independent runs. 
The perturbation probability $q$ and the offline data ratio $r_\text{off}$ are set to $q=0.5$, $r_\text{off}=0$ in the Hopper environment.  
In the HalfCheetah and Ant environments, $q$ is set to $0.1$ and $r_\text{off}$ to $0.1$.
Further evaluation under different values of $r_\text{off}$ is reported in the Appendix~\ref{appdx_hyperparams}.

In the schedule-$q$ setting, training is extended to 3 million steps to ensure policy convergence.
The smoothing factor $\beta$ in Equation~(\ref{equ:adaptive_schedule}) is fixed to $0.9$ for all environments.  
The step-size parameter $\eta$ is set to $1.0$ in Hopper, $0.7$ in HalfCheetah, and $0.3$ in Ant.

\noindent
\textbf{Perturbation Settings}
In addition to adversarial perturbations (see Section~\ref{4.2}), 
fine-tuning is also conducted under two other conditions: normal and random perturbations. 
Formally, the perturbation vector is defined as:
\begin{equation}
\label{equ:perturbations_case}
\vect{\delta} =
    \begin{cases}
    \vect{0} & \text{normal}, \\ 
    \vect{\delta}_r\sim U([-\epsilon,\epsilon]^{N_a}) & \text{random perturbation}, \\
    \vect{\delta}_a\sim \text{DE}(\epsilon) & \text{adversarial perturbation}.
\end{cases}
\end{equation}
In the normal condition, no perturbation is applied ($\vect{\delta}=\vect{0}$).
For the random perturbation condition, $\vect{\delta}_r$ is sampled from a uniform distribution over $[-\epsilon,\epsilon]^{N_a}$, where $\epsilon$ controls the perturbation strength and $N_a$ denotes the action dimension.
In the Hopper-v2 and HalfCheetah-v2 environments, the magnitudes of $\vect{\delta}_r$ and $\vect{\delta}_a$ are fixed at $\epsilon = 0.3$ during both training and evaluation.
In the Ant-v2 environment, the perturbation magnitude is fixed at $\epsilon = 0.5$. 
Adversarial perturbations used during evaluation are newly generated using the differential evolution algorithm and are independent of those used during fine-tuning.
\begin{figure}[t]
    \centering
     \includegraphics[width=1.0\hsize]{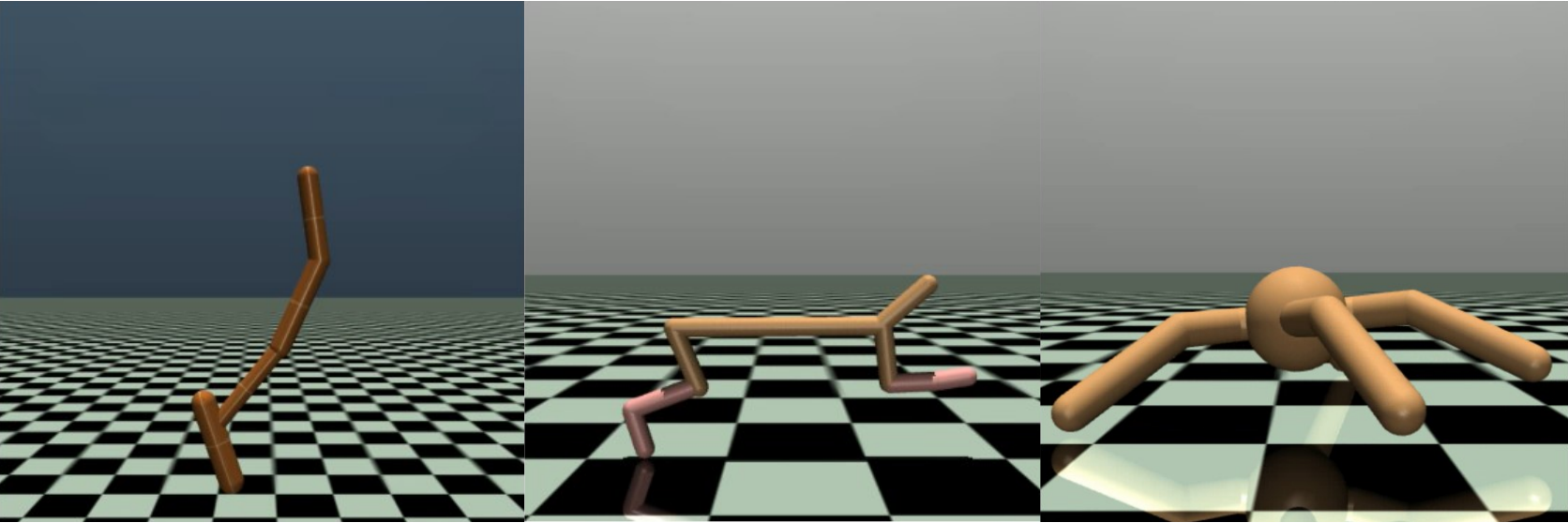}
    \caption{Legged robot environments: Hopper-v2 (left), HalfCheetah-v2 (center), Ant-v2 (right).}
    \label{fig:mujoco}
\end{figure}

\noindent
\textbf{Legged Robot Environments}
We use forward walking tasks in the Hopper-v2, HalfCheetah-v2, and Ant-v2 environments of OpenAI Gym~\cite{OpenAI}. 
These tasks are simulated within the MuJoCo physics engine, as illustrated in Figure~\ref{fig:mujoco}.
The reward functions for each robot are defined as follows: 
\begin{align}
    \label{equ:env_reward}
    r(\vect{s},\vect{a})=
    \begin{cases}
    v_{\text{fwd}}-0.001\left\|\vect{a}\right\|^{2}+1,\ &\text{Hopper-v2}\\
    v_{\text{fwd}}-0.1\left\|\vect{a}\right\|^{2}+1,\  &\text{HalfCheetah-v2}\\
    v_{\text{fwd}}-0.5\|\vect{a}\|^{2}-0.5\cdot10^{-3}\left\|\vect{f}\right\|^{2}+1,\  &\text{Ant-v2}
    \end{cases}
\end{align}
where $v_{\text {fwd }}$ is the forward walking speed and $\vect{f}$ denotes the contact force. 

\noindent
\textbf{Robustness Evaluation}
We evaluate the robustness of fine-tuned policies on a forward walking task in a legged robot environment. Each policy is evaluated under three distinct conditions: normal (i.e., no perturbation), with random perturbations, and with adversarial perturbations.
In the random and the adversarial perturbation condition, the perturbation probability is set to $q=1$.
The evaluation metric is the D4RL-normalized episodic reward over 100 episodes in each condition. 
All experiment results are averaged across the 5 independent runs.
Specifically, the D4RL-normalized episodic reward is calculated as
\begin{equation}
    \label{equ:d4rl_normalize}
    \mathrm{Normalized\ Episodic\ Reward} =
    100 \times
    \frac{\bar{R}_{\mathrm{episode}} - \mathrm{score}_{\mathrm{min}}^{\mathrm{D4RL}}}
    {\mathrm{score}_{\mathrm{max}}^{\mathrm{D4RL}} - \mathrm{score}_{\mathrm{min}}^{\mathrm{D4RL}}},
\end{equation}
where $\mathrm{score}_{\mathrm{min}}^{\mathrm{D4RL}}$ and $\mathrm{score}_{\mathrm{max}}^{\mathrm{D4RL}}$ are predefined for each environment in D4RL and correspond to the performance of a random policy and an expert policy, respectively.

\subsection{Robustness Evaluation of Fine-tuned Models against Baselines}
\label{5.2}
\begin{table}[ht]
    \caption{
    Fine-tuning improves robustness and achieves the highest scores, in contrast to conservative offline methods that drop to sub-zero rewards under adversarial perturbations.
    Values are reported as D4RL-normalized episodic reward ($\text{mean}\pm\text{SE}$). \textit{Offline} results are from~\cite{offline_robust}. \textit{Fully Online} denotes policies trained from scratch under adversarial perturbations. \textit{Fine-tuned} denotes pretrained offline policies further trained under the specified perturbation condition. Bold numbers indicate the best performance for each evaluation condition.
    }
    \centering
    {\begin{tabular}{llrrrrr}
        \toprule
        \multirow{2}{*}{Environments} & \multirow{2}{*}{\makecell{Evaluation\\Conditions}} 
        & \multicolumn{1}{c}{\multirow{2}{*}{Offline}} 
        & \multicolumn{1}{c}{\multirow{2}{*}{\makecell{Fully Online\\(Adversarial)}}} 
        & \multicolumn{3}{c}{Fine-tuned} \\
        \cmidrule(lr){5-7}
        & &  &  & \multicolumn{1}{c}{Normal} 
        & \multicolumn{1}{c}{Random} & \multicolumn{1}{c}{Adversarial} \\
        \midrule
        \multirow{3}{*}{Hopper-v2}
        &Normal     &$\bf{111.1\pm0.1}$&$80.9\pm1.8$&$88.5\pm1.8$&$102.5\pm0.9$&$84.3\pm1.3$ \\ 
        &Random     &$62.9\pm1.9$&$77.2\pm1.8$&$78.5\pm1.8$&$\bf{95.8\pm1.1}$&$89.3\pm1.1$ \\ 
        &Adversarial&$13.7\pm0.1$&$57.0\pm1.5$&$16.7\pm0.1$&$57.4\pm1.2$&$\bf{83.5\pm0.8}$ \\ 
        \midrule
        \multirow{3}{*}{HalfCheetah-v2}
        &Normal     &$\bf{96.6\pm0.2}$&$34.2\pm0.5$&$92.7\pm0.2$&$93.7\pm0.3$&$89.7\pm0.3$ \\ 
        &Random     &$75.0\pm0.8$&$30.5\pm0.5$&$73.4\pm0.6$&$\bf{82.7\pm0.5}$&$73.8\pm0.6$ \\ 
        &Adversarial&$12.1\pm0.4$&$27.6\pm0.4$&$41.0\pm0.7$&$51.7\pm0.8$&$\bf{55.3\pm0.3}$ \\
        \midrule
        \multirow{3}{*}{Ant-v2}
        &Normal     &$119.9\pm1.8$&$34.4\pm0.6$&$127.7\pm1.4$&$\bf{134.2\pm1.1}$&$130.0\pm0.8$ \\ 
        &Random     &$86.6\pm1.9$&$31.8\pm0.6$&$103.4\pm1.6$&$\bf{104.6\pm1.5}$&$99.6\pm1.5$ \\ 
        &Adversarial&$-21.0\pm1.2$&$24.0\pm0.7$&$32.9\pm1.4$&$56.2\pm1.1$&$\bf{91.6\pm0.5}$ \\
        \bottomrule
    \end{tabular}}
    \label{tab:results_ft}
\end{table}
\begin{figure*}[t]
    \centering
    \includegraphics[width=\textwidth]{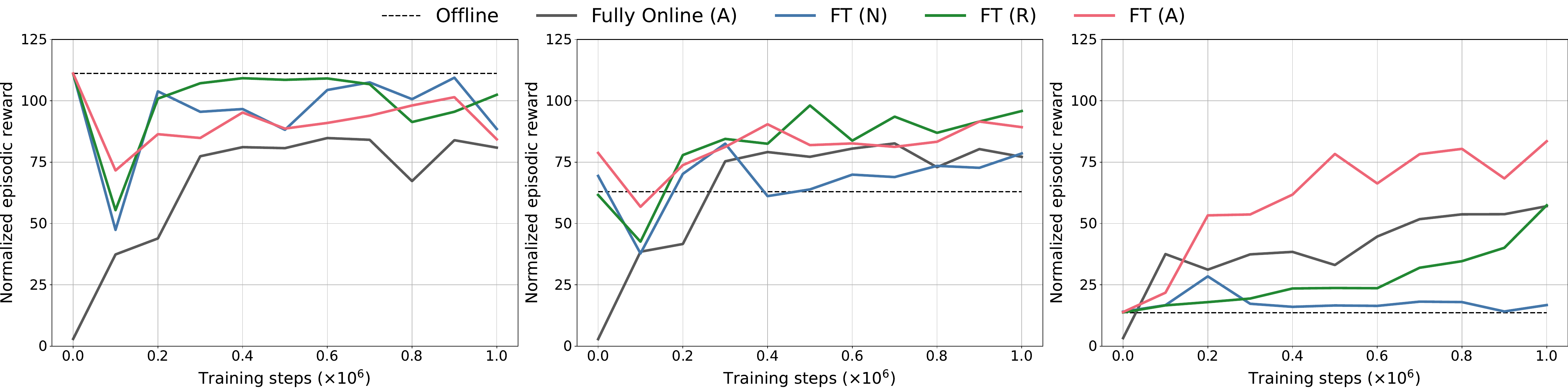}
    (a) Hopper-v2 \par
    \includegraphics[width=\textwidth]{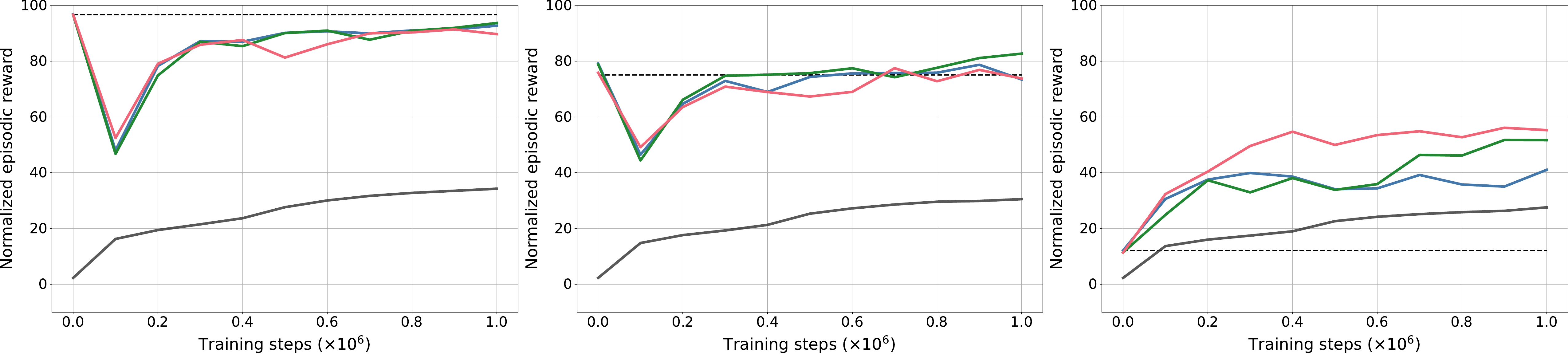}
    (b) HalfCheetah-v2 \par
    \includegraphics[width=\textwidth]{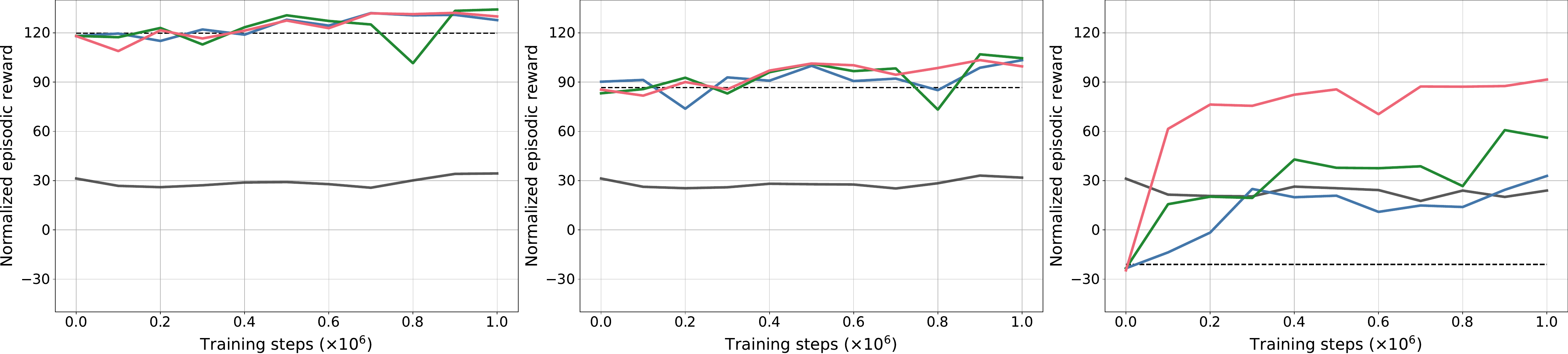}
    (c) Ant-v2 \par
    \caption{
        Performance curves show that fine-tuning accelerates convergence and improves robustness.
        Matching training and evaluation conditions (e.g., FT-A for adversarial evaluation, red curves in right column) yields fastest convergence. Offline baselines (dashed) excel in normal conditions but fail under perturbations. 
        Rows: Hopper-v2, HalfCheetah-v2, Ant-v2. Columns: evaluation under normal, random, and adversarial perturbations. 
        Shaded regions show standard deviation.
    }
    \label{fig:performance_curve}
\end{figure*}
Based on the fine-tuning and perturbation settings defined in Section~\ref{5.1}, we evaluate the proposed adversarial fine-tuning method by comparing it with several baseline and control settings.
Specifically, we compare the proposed Fine-tuned (Adversarial) setting with:
(1) \textit{Offline}: Offline-only training without online interaction,
(2) \textit{Fully Online (Adversarial)}: Training from scratch under adversarial perturbations, and
(3) \textit{Fine-tuned (Normal)} and \textit{Fine-tuned (Random)}: Non-adversarial fine-tuning settings.
\textit{Fine-tuned (Normal)} performs fine-tuning without applying perturbations, and \textit{Fine-tuned (Random)} applies uniformly sampled random perturbations during fine-tuning.
These comparisons allow us to isolate the contribution of offline pretraining and to distinguish adversarial perturbations from non-adversarial perturbation injection.

Table~\ref{tab:results_ft} presents the comparison of these settings across all evaluation conditions.
Values in the table are reported as mean ± standard error (SE), where SE is computed over $500$ evaluation episodes across five independent runs.
Offline-only policies fail under action perturbations due to their lack of exposure to distribution shifts during training. 
For example, in Ant-v2 under adversarial conditions, adversarial fine-tuning achieves a score of $91.6$, compared to $-21.0$ for offline training and $24.0$ for fully online training. A similar trend is observed in Hopper-v2, where the offline score drops to $13.7$ while the fine-tuned score reaches $83.5$. 
These results highlight that adversarially fine-tuned policies are far more robust to action-space perturbations than policies trained solely offline. 
Policies fine-tuned under the normal conditions exhibit severe performance degradation when evaluated under the adversarial perturbation condition (e.g., $16.7$ in Hopper-v2). 
This performance degradation indicates that robustness is not automatically acquired without explicit exposure to perturbed environments during training. 
Robustness is closely tied to the distribution encountered during fine-tuning.
Therefore, performance cannot be maintained when the evaluation condition deviates from the training condition.

Policies fine-tuned under adversarial perturbations retain substantial performance under the random perturbation condition (e.g., $89.3$ in Hopper-v2). 
This retained performance suggests that learning under perturbed environments does not result in over-specialization to a specific perturbation pattern. 
Learning under perturbed environments also improves more fundamental control properties, such as action-output stability and compensatory control ability.
These improvements contribute to a certain degree of general robustness.
However, the best performance is achieved when the fine-tuning condition matches the evaluation condition.
This dependency implies that the acquired robustness is not fully perturbation-independent. 
The acquired robustness includes perturbation-specific adaptation in addition to the shared stability improvement.

Figure~\ref{fig:performance_curve} shows the learning dynamics under different perturbation scenarios. Fine-tuned policies typically exhibit an initial performance drop but recover quickly, converging faster than fully online training. 
This recovery occurs because adversarial perturbations promote exploration of previously unseen action regions, enabling the policy to learn compensatory behaviors more effectively. When the perturbation condition during fine-tuning matches the evaluation condition, convergence becomes faster and more stable, further supporting the advantage of targeted fine-tuning.

In this study, sample efficiency is evaluated based on the number of online environment interactions required during fine-tuning.

From this perspective, these results demonstrate that adversarial fine-tuning effectively combines the robustness advantages of online training with the sample efficiency of offline pretraining. 
This approach provides a practical solution for deploying RL policies in real-world robotic systems where actuator perturbations are inevitable.
To further examine the algorithmic generality of the proposed framework, we conduct additional experiments using Implicit Q-Learning (IQL)~\cite{IQL} pretraining; detailed results are provided in Appendix~\ref{appdx_IQL}.
We further investigate how different perturbation generation mechanisms influence fine-tuning behavior; details are provided in Appendix~\ref{appdx:pgd_adv}.

\subsection{Improving Adversarial Robustness via Curriculum Fine-tuning}
\label{5.3}
We compare three strategies: fixed perturbation probability ($q_\text{fix}$), linearly increasing curriculum ($q_\text{lin}$), and adaptive curriculum ($q_\text{ada}$) that adjusts $q$ based on policy performance. 
To isolate the effect of adaptive scheduling, we calibrate $q_\text{lin}$'s maximum value $q_\text{max}$ so that both curricula provide equal average perturbation exposure over training.

Table~\ref{tab:results_q_adaptive} compares the three strategies across training durations and evaluation conditions. 
At $1$M steps under adversarial conditions, $q_\text{lin}$ shows inconsistent performance: lower than $q_\text{fix}$ in Hopper-v2 ($30.7$ vs $83.5$) and Ant-v2 ($84.6$ vs $91.6$), but comparable in HalfCheetah-v2 ($58.7$ vs $55.3$). 
In contrast, $q_\text{ada}$ consistently outperforms $q_\text{fix}$ across all three environments (Hopper-v2: $87.9$ vs $83.5$; HalfCheetah-v2: $63.4$ vs $55.3$; Ant-v2: $98.2$ vs $91.6$), demonstrating more reliable early-stage robustness.
By $3$M steps, the two curricula exhibit distinct trade-offs. 
Although $q_\text{lin}$ improves adversarial performance in some environments (Ant-v2: $84.6$ at $1$M to $117.7$ at $3$M), it degrades normal performance (Hopper-v2: $95.1$ to $76.5$; HalfCheetah-v2: $91.7$ to $88.7$; Ant-v2: $125.0$ to $123.7$). 
In contrast, $q_\text{ada}$ maintains or improves normal performance across environments (Hopper-v2: 100.2 to 97.5; HalfCheetah-v2: $89.0$ to $93.0$; Ant-v2: $124.8$ to $130.8$) while achieving comparable adversarial robustness (Ant-v2: $118.5$ vs $117.7$). 

\begin{figure}[t]
    \centering
    \includegraphics[width=\textwidth]{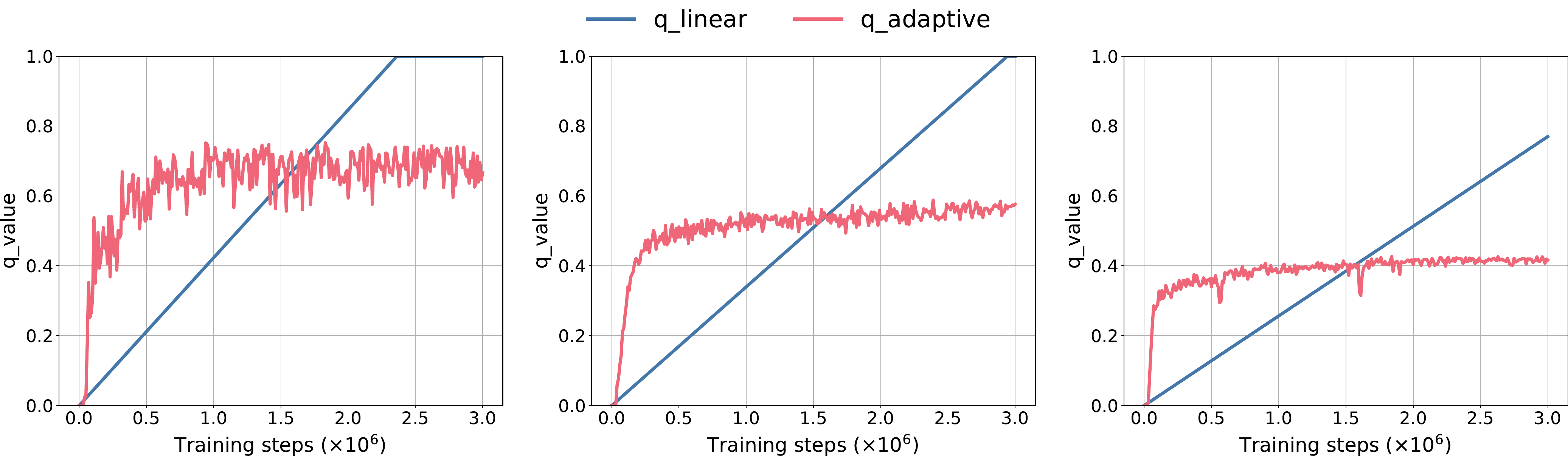}
    \caption{
    The adaptive curriculum ($q_\text{ada}$), which adjusts the perturbation probability based on performance, mitigates the degradation in performance under the normal condition observed under the linear schedule ($q_\text{lin}$) and maintains robust performance across evaluation conditions.
    Results are reported for the fixed-$q$ setting ($q_\text{fix}$), linearly increasing strategy ($q_\text{lin}$), and adaptively adjusted strategy ($q_\text{ada}$) after $1\text{M}$, $2\text{M}$, and $3\text{M}$ training steps.
    Values indicate D4RL-normalized episodic rewards ($\text{mean} \pm \text{standard error}$) under normal, random, and adversarial perturbation conditions.
    }
    \label{fig:q_value_curve}
\end{figure}

Figure~\ref{fig:q_value_curve} shows that under $q_\text{lin}$, the perturbation probability increases monotonically throughout training.
In the linear schedule setting, the decline in the normal performance arises from a distributional imbalance induced by this monotonic increase in perturbation probability. 
As the perturbation probability increases, transitions stored in the replay buffer $\mathcal{R}$ become progressively dominated by data collected under the perturbed environment.
Consequently, gradient updates are biased toward adaptation to noisy conditions, while stable control behaviors required under the normal environments receive relatively less reinforcement. 
This imbalance weakens nominal behaviors acquired during offline pretraining and leads to the observed performance degradation.

In contrast, the adaptive schedule adjusts $q$ according to policy performance.
This adjustment prevents excessive concentration of transitions from the perturbed environment in the replay buffer and mitigates the distributional imbalance.
This demonstrates that performance-based adaptation successfully avoids the stability-robustness trade-off inherent in fixed scheduling. 
Moreover, it suggests that dynamic curriculum mechanisms can offer a principled way to balance robustness and stability in broader reinforcement learning settings.
\begin{table}[t]
    \caption{
    The adaptive curriculum ($q_\text{ada}$), which adjusts based on performance, suppresses the degradation in nominal performance caused by overfitting in the linear curriculum ($q_\text{lin}$) and maintains high performance.
    Results are reported for fixed-$q$ setting ($q_\text{fix}$), linearly increasing strategy ($q_\text{lin}$), and adaptively adjusted strategy ($q_\text{ada}$) after 1M, 2M, and 3M training steps.  
    Values indicate D4RL-normalized episodic rewards under normal, random, and adversarial perturbation conditions.
    }
    \centering
    {\begin{tabular}{llrrrrrrr}
        \toprule
        \multirow{2}{*}{Environments} & \multirow{2}{*}{\makecell{Evaluation\\Conditions}} 
        & \multicolumn{1}{c}{\multirow{2}{*}{$q_\text{fix}$}} 
        & \multicolumn{3}{c}{$q_\text{lin}$} 
        & \multicolumn{3}{c}{$q_\text{ada}$} \\
        \cmidrule(lr){4-6}
        \cmidrule(lr){7-9}
        & & &\multicolumn{1}{c}{$1\text{M}$} 
        & \multicolumn{1}{c}{$2\text{M}$} & \multicolumn{1}{c}{$3\text{M}$} 
        & \multicolumn{1}{c}{$1\text{M}$} 
        & \multicolumn{1}{c}{$2\text{M}$} & \multicolumn{1}{c}{$3\text{M}$} \\
        \midrule
        \multirow{3}{*}{Hopper-v2}
        &Normal     &$ 84.3$&$ 95.1$&$ 86.8$&$ 76.5$&$100.2$&$ 75.0$&$ 97.5$ \\ 
        &Random     &$ 89.3$&$ 71.4$&$ 76.2$&$ 85.6$&$ 92.5$&$ 74.2$&$ 93.8$ \\ 
        &Adversarial&$ 83.5$&$ 30.7$&$ 85.2$&$ 79.1$&$ 87.9$&$ 88.2$&$ 81.5$ \\ 
        \midrule
        \multirow{3}{*}{HalfCheetah-v2}
        &Normal     &$ 89.7$&$ 91.7$&$ 91.9$&$ 88.7$&$ 89.0$&$ 93.8$&$ 93.0$ \\ 
        &Random     &$ 73.8$&$ 74.8$&$ 78.0$&$ 74.0$&$ 70.9$&$ 79.0$&$ 76.6$ \\ 
        &Adversarial&$ 55.3$&$ 58.7$&$ 72.4$&$ 76.4$&$ 63.4$&$ 69.2$&$ 75.1$ \\
        \midrule
        \multirow{3}{*}{Ant-v2}
        &Normal     &$130.0$&$125.0$&$130.5$&$123.7$&$124.8$&$131.0$&$130.8$ \\ 
        &Random     &$ 99.6$&$ 91.9$&$ 90.3$&$ 98.8$&$ 95.5$&$ 97.3$&$110.1$ \\ 
        &Adversarial&$ 91.6$&$ 84.6$&$104.2$&$117.7$&$ 98.2$&$115.7$&$118.5$ \\
        \bottomrule
    \end{tabular}}
    \label{tab:results_q_adaptive}
\end{table}

\section{Limitations and Future Directions}
\label{6}
This study has three main limitations.
First, this study focuses on static multiplicative action perturbations within each episode. 
These perturbations represent continuous execution errors such as actuator degradation or gain mismatch. 
However, more complex forms of action uncertainty are not addressed. 
For example, action delays and actuator saturation change the timing between actions and state transitions or impose nonlinear constraints. 
Because these effects change the control dynamics, our framework cannot be directly applied without reformulation.
Second, the proposed method addresses action-space (output-level) perturbations, while many prior robust RL approaches~\cite{state_perturbation_1,state_perturbation_3} focus on state-space (input-level) perturbations. 
Since they affect different parts of the model, combining state-robust methods with the proposed action-space adaptation framework is a possible direction for future work.
Finally, a more principled formulation that explicitly models the trade-off between nominal performance and adversarial robustness remains an important direction for future work. 
In the current framework, this balance is handled implicitly and may still depend on the specific task. 
An explicit multi-objective formulation~\cite{morl_1,morl_2} could provide a clearer and more systematic understanding of this balance.

\section{Conclusion}
\label{7}
Conservative offline RL methods ensure stability but conflict with action-space robustness because they prohibit learning from the out-of-distribution samples necessary for adaptation to perturbations. 
This study addressed this incompatibility through an offline-to-online framework with adversarial fine-tuning, where policies pretrained offline are subsequently fine-tuned in perturbed environments. 
Additionally, we introduced an adaptive curriculum that adjusts the perturbation probability based on policy performance to prevent overfitting while maintaining training stability.

Experiments using legged robots demonstrated that our method outperformed both offline-only and fully online baselines and converged faster than training from scratch. 
In particular, the adaptive curriculum alleviated the overfitting observed in simpler linear schedules, which often sacrificed normal performance, thereby achieving an effective trade-off between robustness and stability.

\section*{Acknowledgments}
This work was supported by JSPS KAKENHI Grant Number JP23K24914.

\bibliographystyle{unsrt}  
\bibliography{refs_arxiv}

@article{offline_rl_1,
  title={Offline Reinforcement Learning: Tutorial, Review, and Perspectives on Open Problems},
  author={Levine, Sergey and Kumar, Aviral and Tucker, G. and Fu, Justin},
  archivePrefix={arXiv},
  journal={arXiv:2005.01643},
  year={2020},
}

@article{offline_rl_2,
    author={Figueiredo Prudencio, Rafael and Maximo, Mardos, R. O. A. and Colombini, Esther Luna},
    journal={IEEE Transactions on Neural Networks and Learning Systems}, 
    title={A Survey on Offline Reinforcement Learning: Taxonomy, Review, and Open Problems}, 
    year={2024},
    volume={35},
    number={8},
    pages={10237-10257},
}

@InProceedings{BCQ,
    title={Off-Policy Deep Reinforcement Learning without Exploration},
    author={Fujimoto, Scott and Meger, David and Precup, Doina},
    booktitle={the 36th International Conference on Machine Learning (ICML 2019)},
    year={2019},
    volume={97},
    pages={2052-2062},
}

@inproceedings{TD3+BC,
    author = {Fujimoto, Scott and Gu, Shixiang (Shane)},
    title = {A Minimalist Approach to Offline Reinforcement Learning},
    booktitle = {the 35th Conference on Neural Information Processing Systems (NeurIPS 2021)},
    year = {2021},
    pages = {20132-20145},
    volume = {34},
}

@inproceedings{IQL,
    title={Offline Reinforcement Learning with Implicit Q-Learning},
    author={Kostrikov, Ilya and Nair, Ashvin Nair and Levine, Sergey},
    booktitle={the 10th International Conference on Learning Representations (ICLR 2022)},
    year={2022}
}

@inproceedings{CQL,
    author={Kumar, Aviral and Zhou, Aurick and Tucker, George and Levine, Sergey},
    title={Conservative Q-learning for offline reinforcement learning},
    year={2020},
    booktitle={the 34th Conference on Neural Information Processing Systems (NeurIPS 2020)},
    pages = {1179--1191},
    volume = {33},
}

@inproceedings{Fisher-BRC,
    title={Offline Reinforcement Learning with Fisher Divergence Critic Regularization},
    author={Kostrikov, Ilya and Tompson, Jonathan and Fergus, Rob and Nachum, Ofir},
    booktitle={the 38 th International Conference on Machine
Learning (ICML 2021)},
    year={2021},
    volume={139},
    pages={5774-5783}
}

@article{offline_rl_hc,
    author = {Emerson, Harry and Guy, Matthew and McConville, Ryan},
    title = {Offline reinforcement learning for safer blood glucose control in people with type 1 diabetes},
    journal = {Journal of Biomedical Informatics},
    volume = {142},
    pages = {104376},
    year = {2023},
    issn = {1532-0464},
}

@inproceedings{offline_rl_em,
    title = {DeepThermal: Combustion Optimization for Thermal Power Generating Units Using Offline Reinforcement Learning},
    author = {Zhan, Xianyuan and Xu, Haoran and Zhang, Yue and Zhu, Xiangyu and Yin, Honglei and Zheng, Yu},
    year = {2022},
    month = {06},
    pages = {4680-4688},
    volume = {36},
    booktitle = {the 36th AAAI Conference on Artificial Intelligence (AAAI 2022)},
}

@inproceedings{offline_rl_rc,
    title={Benchmarking Offline Reinforcement Learning on Real-Robot Hardware},
    author={Nico G{\"u}rtler and others},
    booktitle={the 11th International Conference on Learning Representations (ICLR 2023)},
    year={2023}
}

@inproceedings{RORL,
    author = {Yang, Rui and Bai, Chenjia and Ma, Xiaoteng and Wang, Zhaoran and Zhang, Chongjie and Han, Lei},
    booktitle = {the 36th Conference on Neural Information Processing System (NeurIPS 2022)},
    pages = {23851-23866},
    title = {{RORL}: Robust Offline Reinforcement Learning via Conservative Smoothing},
    volume = {35},
    year = {2022},
}

@inproceedings{MICRO,
  title     = {{MICRO}: Model-Based Offline Reinforcement Learning with a Conservative Bellman Operator},
  author    = {Liu, Xiao-Yin and Zhou, Xiao-Hu and Li, Guotao and Li, Hao and Gui, Mei-Jiang and Xiang, Tian-Yu and Huang, De-Xing and Hou, Zeng-Guang},
  booktitle = {the 33rd International Joint Conference on Artificial Intelligence (IJCAI 2024)},
  pages={4587--4595},
  year={2024},
}

@InProceedings{robust_testtime_5,
    author={Nguyen, Thanh and Luu, Tung M. and Ton, Tri and Yoo, Chang D.},
    title={Towards Robust Policy: Enhancing Offline Reinforcement Learning with Adversarial Attacks and Defenses},
    booktitle={Pattern Recognition and Artificial Intelligence},
    year={2025},

    pages={310--324},
}

@misc{AWAC,
    title={AWAC: Accelerating Online Reinforcement Learning with Offline Datasets}, 
    author={Nair, Ashvin and Gupta, Abhishek and Dalal, Murtaza and Levine, Sergey},
    year={2021},
    journal={arXiv:2006.09359},
    archivePrefix={arXiv},
    primaryClass={cs.LG},
}

@inproceedings{BR,
    title={Offline-to-Online Reinforcement Learning via Balanced Replay and Pessimistic Q-Ensemble},
    author={Lee, Seunghyun and Seo, Younggyo and Lee, Kimin and Abbeel, Pieter and Shin, Jinwoo},
    booktitle={5th Annual Conference on Robot Learning (CoRL 2021)},
    year={2021},
}

@inproceedings{APL,
    title={Adaptive Policy Learning for Offline-to-Online Reinforcement Learning}, 
    author={Zheng, Han and Luo, Xufang and Wei, Pengfei and Song, Xuan and Li, Dongsheng and Jiang, Jing},
    year = {2023},
    month = {06},
    pages = {11372-11380},
    volume = {37},
    booktitle = {the 37th AAAI Conference on Artificial Intelligence (AAAI 2023)},
}

@inproceedings{adaptive_bc,
    title={Adaptive Behavior Cloning Regularization for Stable Offline-to-Online Reinforcement Learning}, 
    author={Zhao, Yi and Boney, Rinu and Ilin, Alexander and Kannala, Juho and Pajarinen, Joni},
    booktitle={the 30th European Symposium on Artificial Neural Networks, Computational Intelligence and
Machine Learning (ESANN 2022)},
    year={2022},
    pages = {545--550},
}

@article{OEMA,
    author={Guo, Siyuan and Zou, Lixin and Chen, Hechang and Qu, Bohao and Chi, Haotian and Yu, Philip S. and Chang, Yi},
     journal={IEEE Transactions on Knowledge and Data Engineering}, 
      title={Sample Efficient Offline-to-Online Reinforcement Learning}, 
     year={2024},
      volume={36},
      number={3},
     pages={1299-1310},
}

@inproceedings{PEX,
    title={Policy Expansion for Bridging Offline-to-Online Reinforcement Learning},
    author={Zhang, Haichao and Xu Wei and Yu, Haonan},
    booktitle={the 11th International Conference on Learning Representations (ICLR 2023)},
    year={2023},
}

@inproceedings{FamO2O,
    title={Train Once, Get a Family: State-Adaptive Balances for Offline-to-Online Reinforcement Learning},
    author={Wang, Shenzhi and Yang, Qisen and Gao, Jiawei and Lin, Matthieu Gaetan and CHEN, HAO and Wu, Liwei and Jia, Ning and Song, Shiji and Huang, Gao},
    booktitle={the 37th Conference on Neural Information Processing Systems (NeurIPS 2023)},
    year={2023},
}

@inproceedings{SAC,
    title={Soft Actor-Critic: Off-Policy Maximum Entropy Deep Reinforcement Learning with a Stochastic Actor}, 
    author={Haarnoja, Tuomas and Zhou, Aurick and Abbeel, Pieter and Levine, Sergey},
    year={2018},
    booktitle ={the 35th International Conference on Machine Learning (ICML 2018)},
    volume={80},
    pages={1861-1870}
}

@inproceedings{TD3,
    author = {Fujimoto, Scott and Hoof, Herke and Meger, Dave},
    booktitle = {the 35 th International Conference on Machine Learning (ICML 2018)},
    pages = {1582-1591},
    title = {Addressing Function Approximation Error in Actor-Critic Methods},
    year = {2018},
}

@INPROCEEDINGS{domain_random_1,
    author={Tobin, Josh and others},
    booktitle={2017 IEEE/RSJ International Conference on Intelligent Robots and Systems (IROS 2017)}, 
    title={Domain randomization for transferring deep neural networks from simulation to the real world}, 
    year={2017},
    volume={},
    number={},
    pages={23-30},
}

@inproceedings{domain_random_2,
    title={Closing the Sim-to-Real Loop: Adapting Simulation Randomization with Real World Experience},
    author={Chebotar, Yevgen and others},
    booktitle={2019 IEEE International Conference on Robotics and Automation (ICRA 2019)},
    year={2018},
    pages={8973-8979}
}

@article{domain_random_3,
    title = {Cyclic policy distillation: Sample-efficient sim-to-real reinforcement learning with domain randomization},
    journal = {Robotics and Autonomous Systems},
    volume = {165},
    pages = {104425},
    year = {2023},
    issn = {0921-8890},
    author = {Kadokawa, Yuki and Zhu, Lingwei and Tsurumine, Yoshihisa and Matsubara, Takamitsu},
}

@inproceedings{adv_trng_1,
    title={Robust Adversarial Reinforcement Learning},
    author={Lerrel Pinto and James Davidson and Rahul Sukthankar and Abhinav Kumar Gupta},
    booktitle={the 34th International Conference on Machine Learning (ICML 2017)},
    volume={70},
    pages={2817-2826},
  year={2017}
}

@inproceedings{adv_trng_2,
    title={Action Robust Reinforcement Learning and Applications in Continuous Control},
    author={Tessler, Chen and Efroni, Yonathan and Mannor, Shie},
    booktitle={the 36th International Conference on Machine Learning (ICML 2019)},
    volume={97},
    pages={6215-6224},
    year={2019},
}

@inproceedings{adv_trng_3,
    title={Robust Policy Gradient against Strong Data Corruption},
    author={Zhang, Xuezhou and Chen, Yiding and Zhu, Xiaojin and Sun, Wen},
    booktitle={the 38th International Conference on Machine Learning (ICML 2021)}, 
    year={2021},
    volume={139},
    pages={12391--12401}
}

@INPROCEEDINGS{adv_trng_4,
    author={Sheng, Junru and Zhai, Peng and Dong, Zhiyan and Kang, Xiaoyang and Chen, Chixiao and Zhang, Lihua},
    booktitle={2022 International Joint Conference on Neural Networks (IJCNN 2022)}, 
    title={Curriculum Adversarial Training for Robust Reinforcement Learning}, 
    year={2022},
    volume={},
    number={},
    pages={1-8},
}

@INPROCEEDINGS{adv_trng_5,
    author={Zheng, Xiang and Ma, Xingjun and Wang, Shengjie and Wang, Xinyu and Shen, Chao and Wang, Cong},
    booktitle={the 54th Annual IEEE/IFIP International Conference on Dependable Systems and Networks (DSN 2024)}, 
    title={Toward Evaluating Robustness of Reinforcement Learning with Adversarial Policy}, 
    year={2024},
    volume={},
    number={},
    pages={288-301},
}

@article{offline_robust,
      title={Robustness Evaluation of Offline Reinforcement Learning for Robot Control Against Action Perturbations}, 
      author={Ayabe, Shingo and Otomo, Takuto and Kera, Hiroshi and Kawamoto, Kazuhiko},
      year={2025},
      journal={International Journal of Advanced Robotic Systems},
}

@article{OpenAI,
      title={OpenAI Gym}, 
      author={Brockman, Greg and Cheung, Vicki and Pettersson , Ludwig and Schneider, Jonas and Schulman, John and Tang, Jie and Zaremba, Wojciech},
      year={2016},
      archivePrefix={arXiv},
      journal={arXiv:1606.01540},
      primaryClass={cs.LG}
}

@inproceedings{o2022_adv,
    title={Adversarial joint attacks on legged robots},
    author={Otomo, Takuto and Kera, Hiroshi and Kawamoto, Kazuhiko},
    booktitle={2022 IEEE International Conference on Systems, Man, and Cybernetics},
    year={2022},
    pages={676-681}
}

@article{d3rlpy,
    author  = {Seno, Takuma and Imai, Michita},
    title   = {d3rlpy: An Offline Deep Reinforcement Learning Library},
    journal = {Journal of Machine Learning Research},
    year    = {2022},
    volume  = {23},
    pages   = {1-20}
}

@article{D4RL,
    title={{D4RL}: Datasets for Deep Data-Driven Reinforcement Learning},
    author={Fu, Justin and Kumar, Aviral and Nachum, Ofir and Tucker, George and Levine, Sergey},
    year={2020},
    archivePrefix={arXiv },
    journal={arXiv:2004.07219},
    primaryClass={cs.LG}
}

@inproceedings{state_perturbation_1,
    author = {Zhang, Huan and Chen, Hongge and Xiao, Chaowei and Li, Bo and Liu, Mingyan and Boning, Duane and Hsieh, Cho-Jui},
    title = {Robust deep reinforcement learning against adversarial perturbations on state observations},
    booktitle = {the 34th Conference on Neural Information Processing Systems (NeurIPS 2020)},
    pages = {21024--21037},
    volume = {33},
    year = {2020}
}

@inproceedings{state_perturbation_2,
    title={Who Is the Strongest Enemy? Towards Optimal and Efficient Evasion Attacks in Deep {RL}},
    author={Sun, Yanchao and Zheng, Ruijie and Liang, Yongyuan and Huang, Furong},
    booktitle={the 10th International Conference on Learning Representations (ICLR 2022)},
    year={2022},
}

@inproceedings{state_perturbation_3,
    title={Robust Reinforcement Learning on State Observations with Learned Optimal Adversary},
    author={Huan Zhang and Hongge Chen and Duane S Boning and Cho-Jui Hsieh},
    booktitle={the 9th International Conference on Learning Representations (ICLR 2021)},
    year={2021},
}

@INPROCEEDINGS{action_perturbation_1,
    author={Liu, Qianmei and Kuang, Yufei and Wang, Jie},
    booktitle={2024 International Joint Conference on Neural Networks (IJCNN 2024)}, 
    title={Robust Deep Reinforcement Learning with Adaptive Adversarial Perturbations in Action Space}, 
    year={2024},
    volume={},
    number={},
    pages={1-8},
}

@inproceedings{morl_1,
    author={Turchetta, Matteo and Krause, Andreas and Trimpe, Sebastian},
    booktitle={2020 IEEE International Conference on Robotics and Automation (ICRA 2020)}, 
    title={Robust Model-free Reinforcement Learning with Multi-objective Bayesian Optimization}, 
    year={2020},
    volume={},
    number={},
    pages={10702--10708},
}

@inproceedings{morl_2,
    title={{PD}-{MORL}: Preference-Driven Multi-Objective Reinforcement Learning Algorithm},
    author={Toygun Basaklar and Suat Gumussoy and Umit Ogras},
    booktitle={The 11th International Conference on Learning Representations (ICLR 2023)},
    year={2023},
}

@article{robust_cross_1,
    author={Xu, Yonghao and Du, Bo and Zhang, Liangpei},
    journal={IEEE Transactions on Geoscience and Remote Sensing}, 
    title={Assessing the Threat of Adversarial Examples on Deep Neural Networks for Remote Sensing Scene Classification: Attacks and Defenses}, 
    year={2021},
    volume={59},
    number={2},
    pages={1604--1617},  
}

@article{robust_cross_2,
    author={Gibert, Daniel and Zizzo, Giulio and Le, Quan and Planes, Jordi},
    journal={IEEE Access}, 
    title={Adversarial Robustness of Deep Learning-Based Malware Detectors via (De)Randomized Smoothing}, 
    year={2024},
    volume={12},
    number={},
    pages={61152--61162},
}

@inproceedings{emsemble_1,
    author = {Chua, Kurtland and Calandra, Roberto and McAllister, Rowan and Levine, Sergey},
    title = {Deep reinforcement learning in a handful of trials using probabilistic dynamics models},
    year = {2018},
    booktitle = {the 32nd International Conference on Neural Information Processing Systems (NeurIPS 2018)},
    pages = {4759–-4770},
    numpages = {12},
}

@misc{emsemble_2,
    title={DEFT: Diverse Ensembles for Fast Transfer in Reinforcement Learning}, 
    author={Simeon Adebola and Satvik Sharma and Kaushik Shivakumar},
    year={2022},
    journal={arXiv:2209.12412},
    archivePrefix={arXiv},
    primaryClass={cs.LG},
}

@article{emsemble_3,
    title = {Ensemble-based Deep Reinforcement Learning for robust cooperative wind farm control},
    author = {Binghao He and Huan Zhao and Gaoqi Liang and Junhua Zhao and Jing Qiu and Zhao Yang Dong},
    journal = {International Journal of Electrical Power \& Energy Systems},
    volume = {143},
    pages = {108406},
    year = {2022},
}

@inproceedings{pgd,
    title={Towards Deep Learning Models Resistant to Adversarial Attacks},
    author={Aleksander Madry and Aleksandar Makelov and Ludwig Schmidt and Dimitris Tsipras and Adrian Vladu},
    booktitle={the 6th International Conference on Learning Representations (ICLR 2018)},
    year={2018},
}

\clearpage
\appendix
\section{Theoretical Analysis: Safety of Curriculum Progression}
\label{appdx_theory}
This section provides a theoretical justification for the proposed adaptive curriculum for offline-to-online RL.
Unlike conservative offline RL analyses that aim to lower-bound value estimates, we focus on the \emph{safety of curriculum progression} when the perturbation probability $q \in [0,1]$ changes over training.
Our goal is to reach high performance under severe perturbations (e.g., $q=1$), while avoiding collapse of nominal behavior during the progression.

\subsection*{\textit{Setup.}}
Let $J(\pi,q)$ denote the expected discounted return of policy $\pi$ when perturbations are applied stochastically with probability $q$.
At curriculum step $n$, we (i) improve the policy under the current perturbation level $q_n$, obtaining $\pi_{n+1}$, and then (ii) update the curriculum to $q_{n+1}$.

\subsection*{\textit{Assumptions.}}
\textbf{(A1) Lipschitz continuity in $q$.}
For any policy $\pi$, $J(\pi,q)$ is $L$-Lipschitz with respect to $q$:
\begin{equation}
    |J(\pi,q_1)-J(\pi,q_2)| \le L|q_1-q_2|, \qquad \forall q_1,q_2\in[0,1],
\end{equation}
where $L>0$ captures the effective sensitivity of performance to increasing perturbation frequency.

\noindent
\textbf{(A2) Non-negative expected improvement at fixed $q$.}
Policy optimization under a fixed perturbation level yields
\begin{equation}
    \Delta J_n := \mathbb{E}\left[J(\pi_{n+1},q_n)-J(\pi_n,q_n)\right] \geq 0.
\end{equation}

\subsection*{\textit{A sufficient condition for safe progression.}}
The following lemma formalizes when increasing difficulty (changing $q$) is safe.

\begin{lemma}[Safe curriculum step]
\label{lem:safe_step}
Under (A1)--(A2), the expected return after policy improvement and curriculum update satisfies
\begin{equation}
    \mathbb{E} \big[J(\pi_{n+1},q_{n+1})\big] \geq
    \mathbb{E} \big[J(\pi_n,q_n)\big]
    + \Delta J_n - L\,\mathbb{E}\big[|q_{n+1}-q_n|\big].
\end{equation}
In particular, if the curriculum step satisfies:
\begin{equation}
    \mathbb{E} \big[|q_{n+1}-q_n|\big] \leq \Delta J_n/L,
\end{equation}
then $\mathbb{E}[J(\pi_{n+1},q_{n+1})] \geq \mathbb{E}[J(\pi_n,q_n)]$, i.e., the progression is safe in expectation.
\end{lemma}

\begin{proof}[Proof sketch]
By (A1), $J(\pi_{n+1},q_{n+1}) \geq J(\pi_{n+1},q_n) - L|q_{n+1}-q_n|$.
Taking expectation and using the definition of $\Delta J_n$ yields the claim.
\end{proof}

\subsection*{\textit{Connection to the proposed adaptive update.}}
Our curriculum update changes $q$ proportionally to an observed performance improvement signal.
Concretely, we use an EMA-smoothed normalized score $\hat{s}_n$:
\begin{equation}
    \hat{s}_n = (1-\beta)\hat{s}_{n-1} + \beta\,\tilde{s}_n,
\end{equation}
and update $q$ with a bounded step size:
\begin{equation}
    q_{n+1} = \Pi_{[0,1]}\big(q_n + \eta \, (\hat{s}_n-\hat{s}_{n-1})\big),
\end{equation}
where $\eta>0$ limits the maximum change per curriculum step and $\Pi_{[0,1]}$ is projection.
This form directly enforces a bounded-change schedule, and the EMA parameter $\beta$ reduces the variance of the improvement estimate under noisy episodic returns.
Therefore, the adaptive update is designed to satisfy the sufficient condition in Lemma~\ref{lem:safe_step} more reliably than schedules that increase $q$ independently of learning progress.

\subsection*{\textit{Why linear curricula can degrade nominal performance.}}
A linear schedule increases $q$ regardless of $\Delta J_n$.
When learning progress slows down (small $\Delta J_n$) but $|q_{n+1}-q_n|$ remains large, the shift term $L|q_{n+1}-q_n|$ can dominate the improvement term $\Delta J_n$ in Lemma~\ref{lem:safe_step}, leading to performance degradation and instability.

\subsection*{\textit{Interpreting task-dependent step sizes.}}
Empirically, we observe that tasks requiring more complex coordinated adaptation tend to prefer smaller curriculum steps $\eta$.
This observation is consistent with the interpretation that such tasks have a larger effective sensitivity $L$, though estimating $L$ precisely is an important direction for future work.

\section{Hyperparameter Analysis for Curriculum-based Fine-tuning}
\label{appdx_schedule}
This section analyzes the hyperparameters for the two curriculum-based fine-tuning strategies: the linear and adaptive curricula.

\begin{figure}[t]
    \centering
    \includegraphics[width=\textwidth]{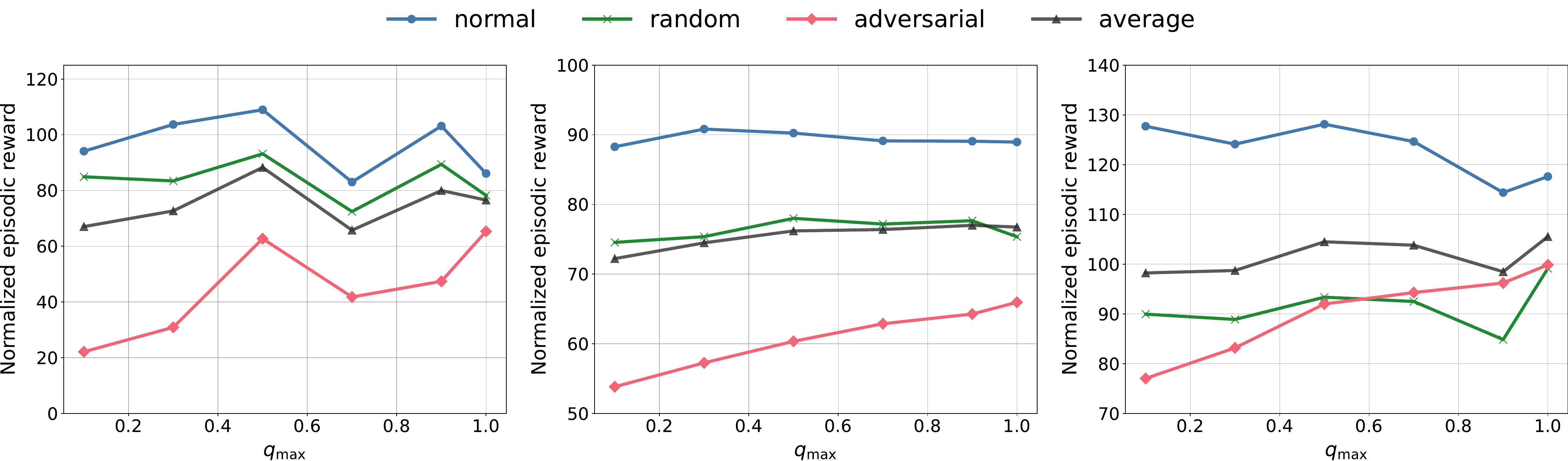}
    \caption{
        Final performance with varying $q_\text{max}$ in Hopper, HalfCheetah, and Ant (left to right). Blue, green, and red denote normal, random, and adversarial conditions; black denotes the average.
    }
    \label{fig:q_max}
\end{figure}

For the linear strategy, the hyperparameter is the maximum perturbation probability $q_\text{max}$.
This value determines the scale of the constant step size $c$ in Equation~(\ref{equ:linear_curriculum}).
Fine-tuning is conducted under adversarial perturbation while varying $q_\text{max}\in\{0.1,0.3,0.5,0.7,0.9,1.0\}$.
Figure~\ref{fig:q_max} shows that performance under adversarial evaluation improves as $q_\text{max}$ increases in all environments, while performance under the normal condition decreases, revealing a trade-off.
This arises because, in the linear strategy, $q$ increases irrespective of policy performance, which induces a bias toward highly perturbed transitions and may result in overfitting to the adversarial condition.

In the adaptive strategy, two hyperparameters are considered: the smoothing factor $\beta$ of the EMA and the step-size parameter $\eta$ that controls the update magnitude.
This section analyzes how these parameters affect the final performance, and the results provide the basis for selecting the values adopted in this study.

\begin{figure}[H]
    \centering
    \includegraphics[width=0.9\textwidth]{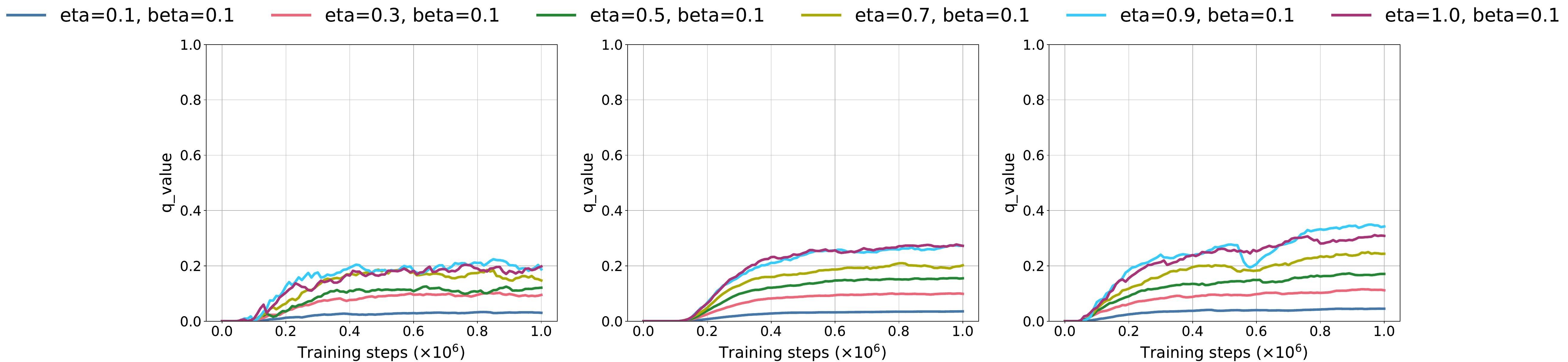}
    \includegraphics[width=0.9\textwidth]{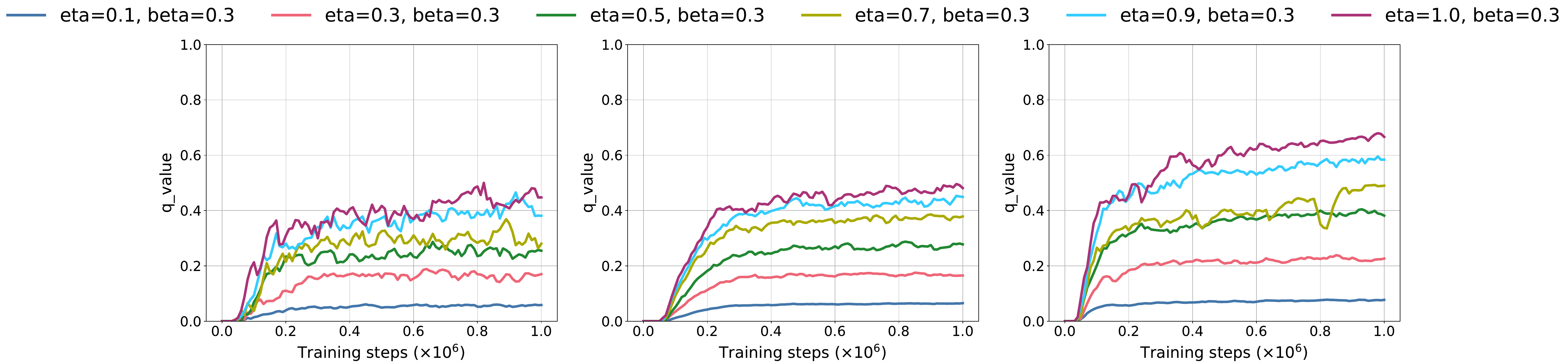}
    \includegraphics[width=0.9\textwidth]{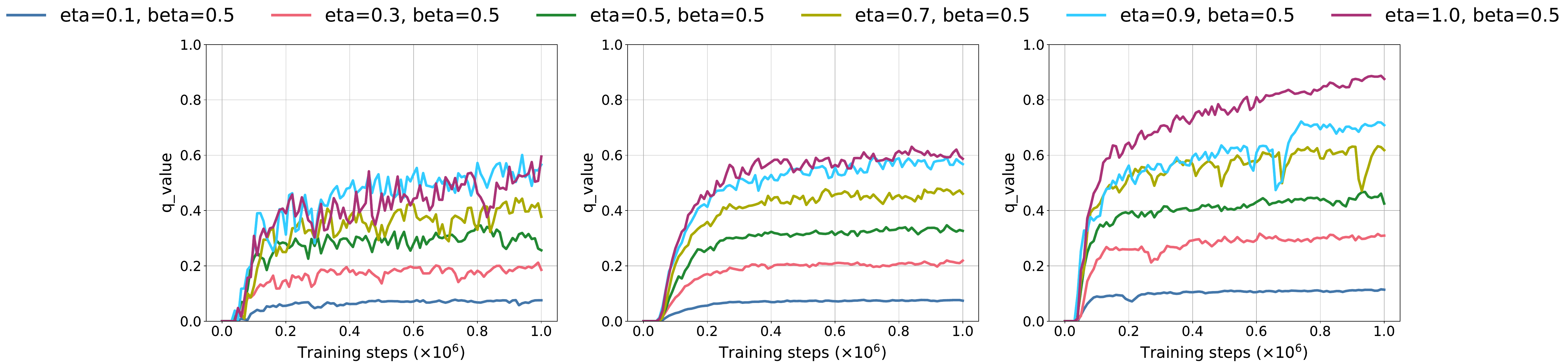}
    \includegraphics[width=0.9\textwidth]{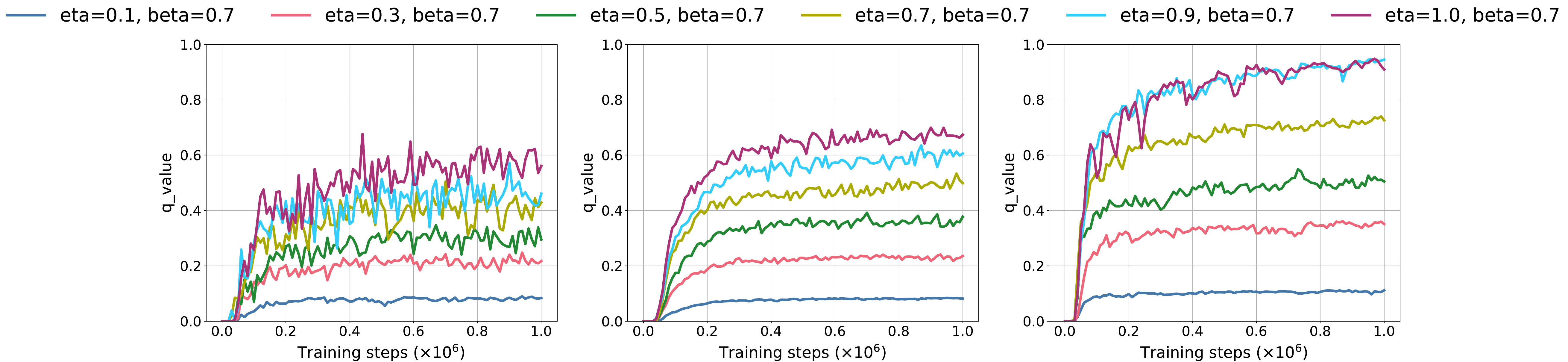}
    \includegraphics[width=0.9\textwidth]{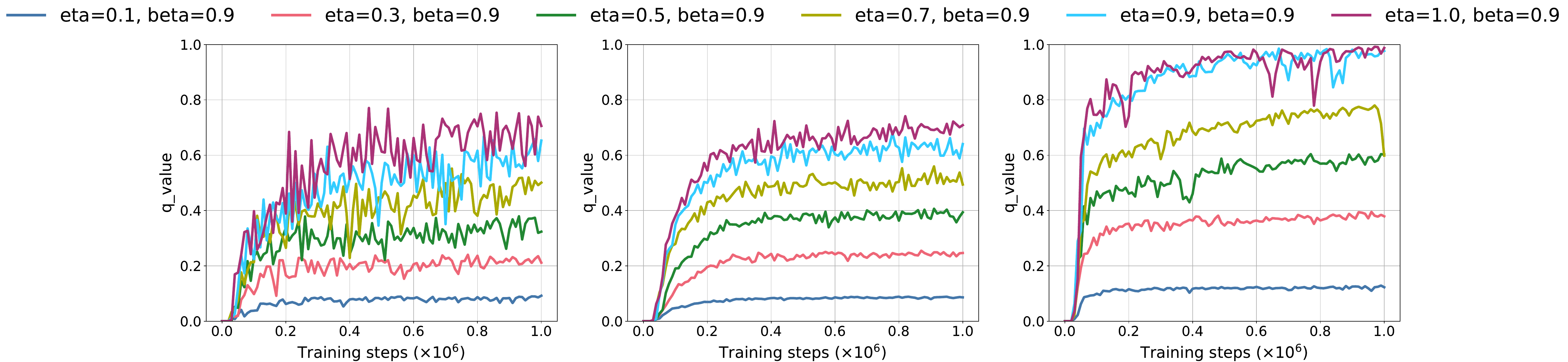}
    \includegraphics[width=0.9\textwidth]{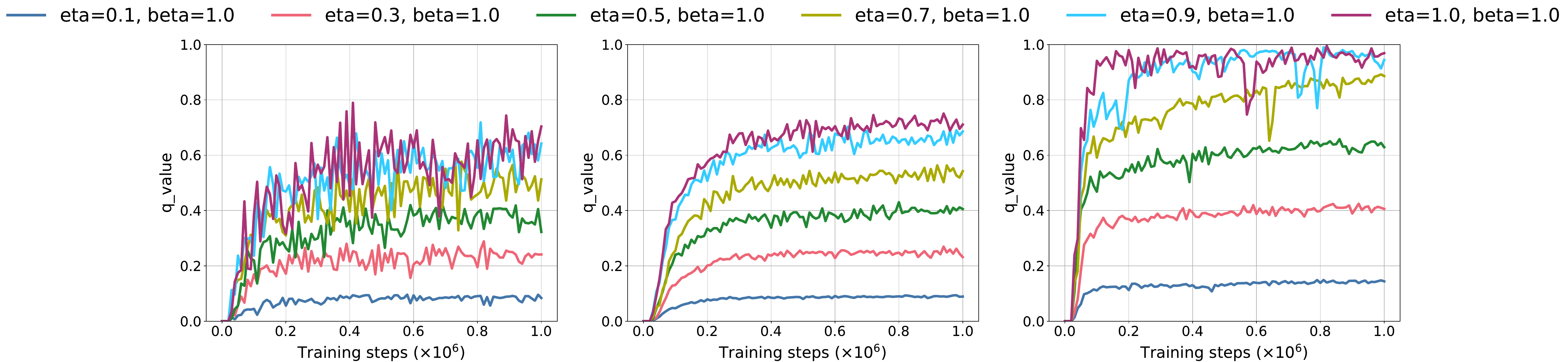}
    \caption{
        Trajectories of the perturbation probability $q$ during fine-tuning for different settings of $\beta$ and $\eta$.
        Panels from left to right correspond to Hopper, HalfCheetah, and Ant environments.  
        Rows from top to bottom correspond to $\beta=0.1,0.3,0.5,0.7,0.9,1.0$.  
    }
    \label{fig:q_curve_beta}
\end{figure}

Figure~\ref{fig:q_curve_beta} shows the trajectory of $q$ for different values of $\beta$ and $\eta$.
The panels from left to right correspond to Hopper, HalfCheetah, and Ant, while rows from top to bottom correspond to $\beta=0.1,0.3,0.5,0.7,0.9,1.0$.
Larger $\beta$ values make $q$ more sensitive to variations in policy performance $R_n$, resulting in greater fluctuations.
For a fixed $\beta$, larger $\eta$ values increase the overall level of $q$, confirming that $q$ is controlled as intended.

\begin{figure}[H]
    \centering
    \includegraphics[width=\textwidth]{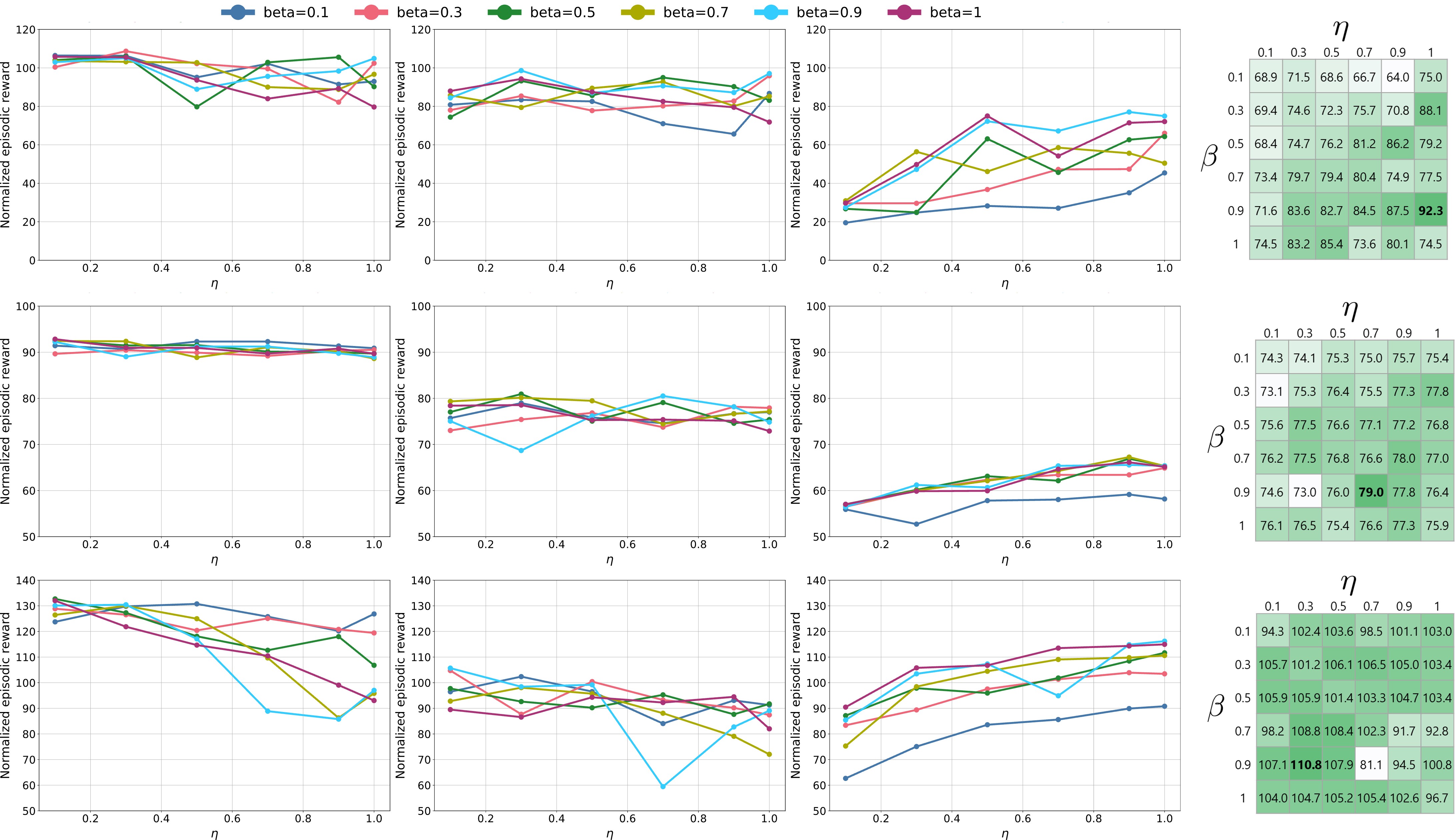}
    \caption{
        Final performance of policies with different setting of $\beta$ and $\eta$.  
        Panels from top to bottom correspond to Hopper, HalfCheetah, and Ant environments.  
        The first three columns plot performance as a function of $\eta$ under normal, random, and adversarial perturbation conditions.  
        The rightmost heatmap summarizes the average performance across three conditions, with darker colors indicating higher values.
    }
    \label{fig:adaptive_performance}
\end{figure}

Figure~\ref{fig:adaptive_performance} summarizes the final performance across environments.
Panels from top to bottom correspond to Hopper, HalfCheetah, and Ant.
The first three columns plot policy performance against $\eta$ under normal, random, and adversarial conditions.
The rightmost heatmap presents the average performance across the three conditions, with darker colors indicating higher values.
Under adversarial perturbations, performance improves as $\eta$ increases.
Under normal and random perturbation conditions, performance varies less with $\beta$ and $\eta$, although Ant shows a stronger trade-off due to larger fluctuations in $q$, similar to the linear strategy.

These results indicate that $\eta$, which controls the update magnitude, and $\beta$, which determines how strongly the update reacts to performance variations, both affect the stability of curriculum progression. In particular, the tendency for larger $\beta$ values to increase fluctuations in $q$ suggests that the update responds more strongly even to small changes in performance. 
This observation is consistent with our theoretical analysis, which highlights the importance of controlling the update magnitude.
The combination of $\beta$ and $\eta$ that maximizes the average performance across the three conditions was selected for the experiments in the main text.


\section{Analysis on Hyperparameters}
\label{appdx_hyperparams}
\begin{figure}[t]
    \centering
    \includegraphics[width=\textwidth]{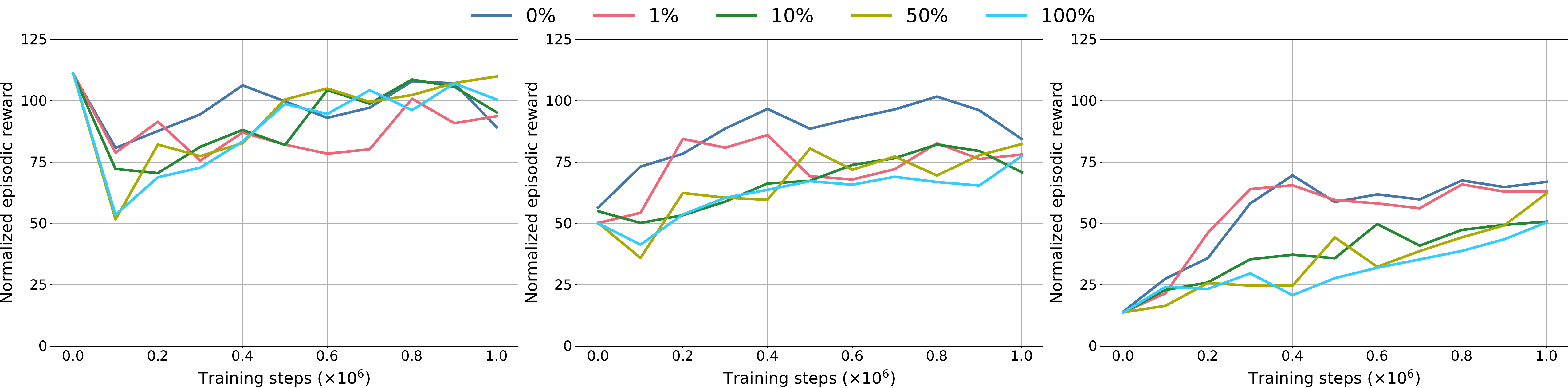}
    (a) Hopper-v2 \par
    \includegraphics[width=\textwidth]{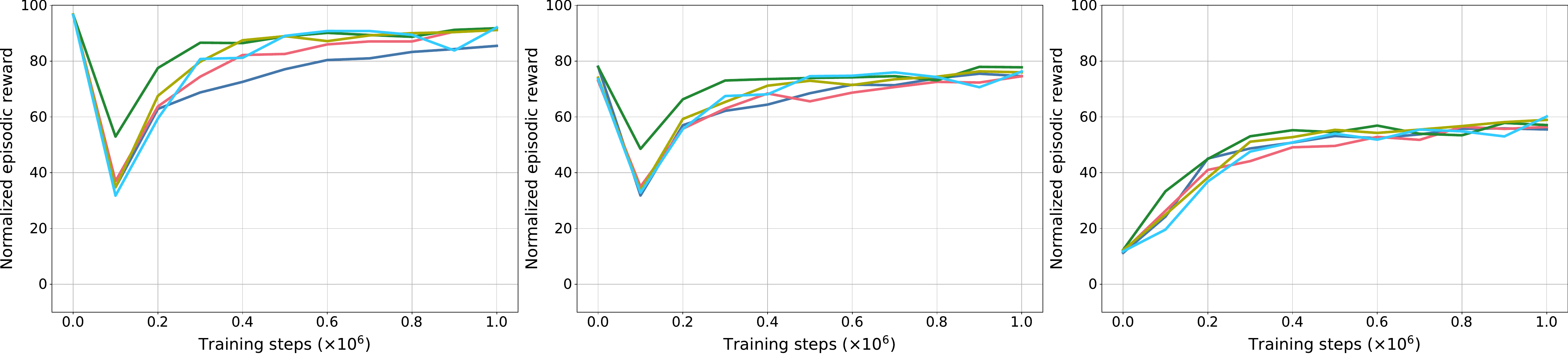}
    (b) HalfCheetah-v2 \par
    \includegraphics[width=\textwidth]{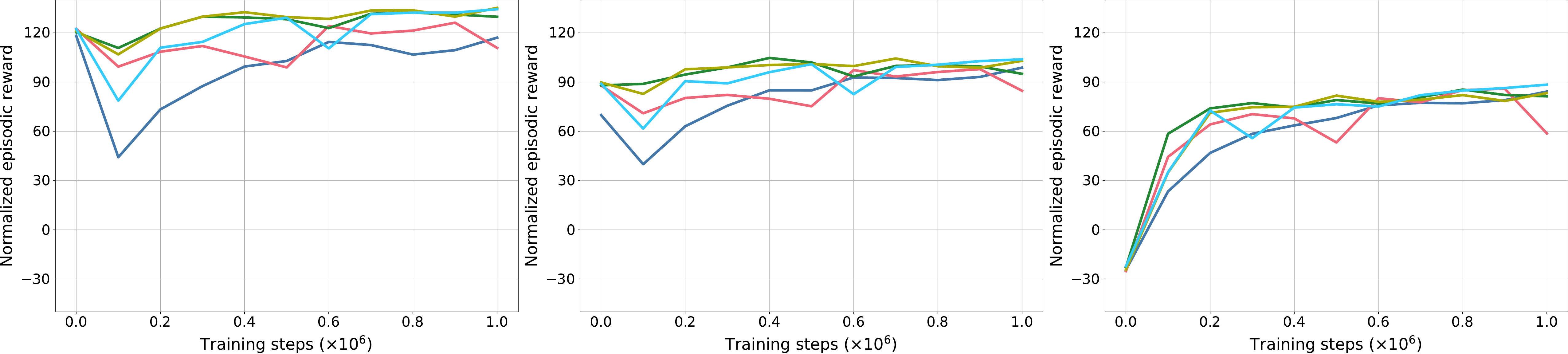}
    (c) Ant-v2 \par
    \caption{
        Performance curves of policies fine-tuned with various $r_\text{off}$ under the adversarial perturbation condition.  
        Subfigures (left to right) correspond to evaluations under the normal, random perturbation, and adversarial perturbation conditions. 
    }
    \label{fig:offdata_adv}
\end{figure}
\begin{figure}[t]
    \centering
    \includegraphics[width=\textwidth]{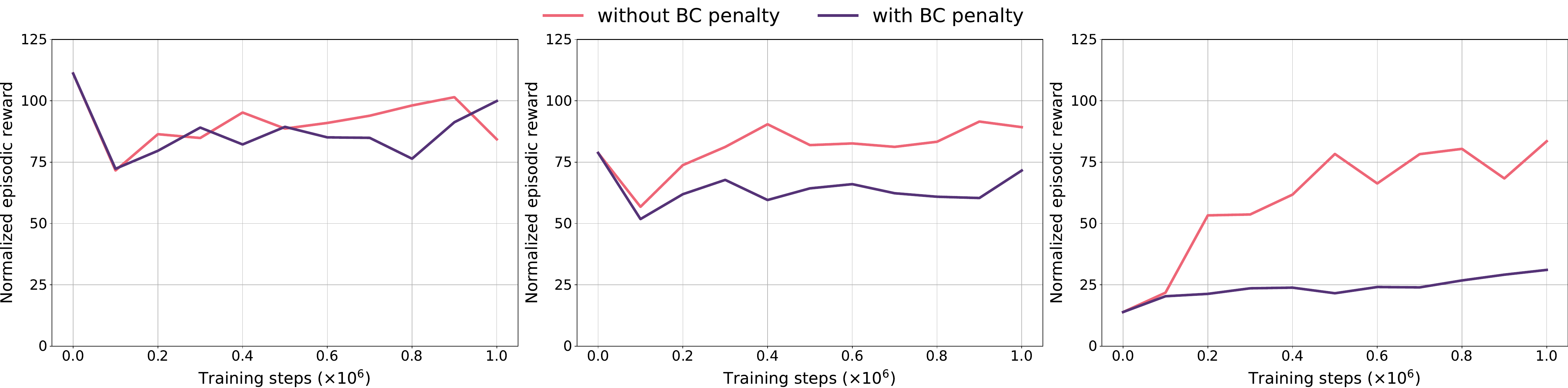}
    (a) Hopper-v2 \par
    \includegraphics[width=\textwidth]{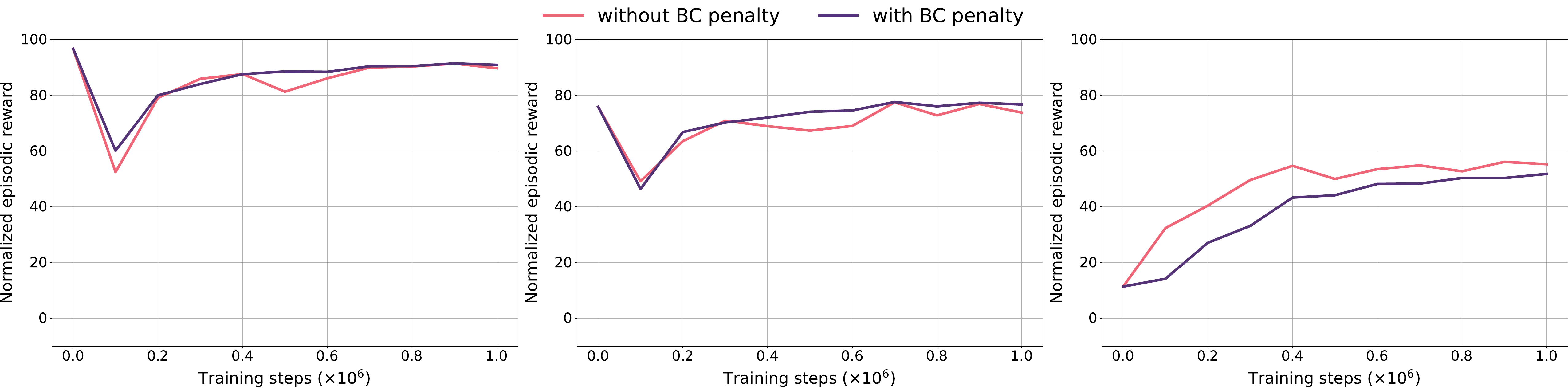}
    (b) HalfCheetah-v2 \par
    \includegraphics[width=\textwidth]{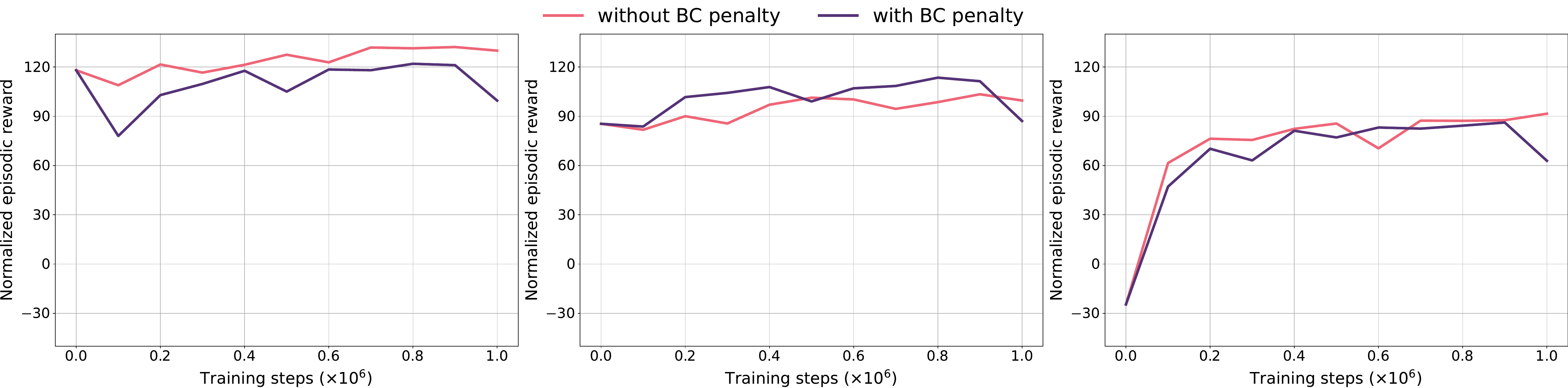}
    (c) Ant-v2 \par
    \caption{
        Performance curves of policies fine-tuned with and without the behavior cloning (BC) penalty under the adversarial perturbation condition.
        Subfigures (from left to right) correspond to evaluations under the normal, random perturbation, and adversarial perturbation conditions.
        }
    \label{fig:bc_adv}
\end{figure}
The effect of the offline data ratio $r_\text{off}$ is evaluated by varying its value among $0$, $0.01$, $0.1$, $0.5$, and $1.0$ under the adversarial fine-tuning condition. 
The corresponding performance curves appear in Figure~\ref{fig:offdata_adv}. 
In HalfCheetah and Ant, offline data affects both convergence speed and final performance. 
In these environments, performance differences become negligible once $r_\text{off}$ exceeds $10\%$. 
In Hopper, smaller values of $r_\text{off}$ yield better performance and produce larger performance fluctuations during adversarial fine-tuning than in HalfCheetah and Ant.
These larger fluctuations indicate a stronger mismatch between clean offline data and the adversarial fine-tuning environment.
Consequently, inserting a large amount of offline data into the replay buffer can slow adaptation.
Based on these observations, environment-specific values of $r_\text{off}$ are selected to achieve faster convergence and higher final performance.

To analyze the effect of the behavior cloning (BC) term in TD3+BC during fine-tuning, additional experiments are conducted under the adversarial condition.
Fine-tuning with and without the BC penalty is compared while keeping all other settings identical.
For each environment, the perturbation probability $q$ and the offline data insertion ratio $r_\text{off}$ follow the same values used in the corresponding experiments without the BC penalty.
This design isolates the impact of the BC term during fine-tuning.

The corresponding performance curves appear in Figure~\ref{fig:bc_adv}.
Retaining the BC penalty slows performance improvement and results in slightly lower final performance across environments.
These results indicate that enforcing the BC penalty during fine-tuning can hinder adaptation to adversarial perturbations.
Accordingly, the BC penalty is removed during fine-tuning in the main experiments.

\section{Additional Experiments with IQL}
\label{appdx_IQL}
To examine whether the proposed method depends on a specific offline reinforcement learning algorithm, we conduct additional experiments using value-based Implicit Q-Learning (IQL)~\cite{IQL}.
While TD3+BC explicitly constrains policy updates through behavior cloning regularization, IQL introduces implicit constraints through expectile-based value estimation and advantage-weighted regression.

We first pretrain the policy using IQL for 3M steps on the expert dataset.
We then perform online fine-tuning for 3M steps using Soft Actor-Critic (SAC)~\cite{sac} under adversarial perturbations.
Adversarial perturbations are generated using Differential Evolution with the same magnitude settings as in the main experiments.
During fine-tuning, we apply the adaptive perturbation probability $q$ proposed in Section~\ref{4.3}
All hyperparameters follow the experimental settings described in Section~\ref{5.1}.

Table~\ref{tab:results_q_adaptive_iql} reports performance under normal, random, and adversarial evaluation conditions.
All results represent the mean D4RL-normalized episodic reward across 5 independent runs.
The \textit{Offline} column corresponds to the pretrained IQL policy without online fine-tuning.
The pretrained performance is comparable to, or slightly lower than, that of the TD3+BC pretrained policy used in the main experiments.
Therefore, the performance gains observed after fine-tuning cannot be attributed to differences in initial performance.

Across all environments, adversarial fine-tuning substantially improves robustness under adversarial evaluation compared to the pretrained policy.
In HalfCheetah-v2 and Ant-v2, nominal performance remains stable while adversarial robustness increases significantly.
In Hopper-v2, adversarial robustness improves during fine-tuning, although nominal performance shows a slight decrease compared to the pretrained policy.
These results indicate that the stability--robustness trade-off is environment-dependent.
This trade-off is particularly pronounced in Hopper-v2, where precise balance control increases sensitivity to robustness-oriented updates under the normal condition.
Taken together, these results demonstrate that the proposed framework generalized beyond policy-constraint-based offline RL methods and remains effective when combined with value-based pretraining.
\begin{table}[H]
    \caption{
    Performance of policies pretrained with IQL and fine-tuned with SAC under adversarial perturbations using the adaptive $q$ schedule.
    \textit{Offline} denotes the pretrained policy without online fine-tuning.
    Columns $1\text{M}$, $2\text{M}$, and $3\text{M}$ report performance after the corresponding numbers of fine-tuning steps.
    Values indicate D4RL-normalized episodic reward ($\text{mean} \pm \text{standard error}$).
    }
    \centering
    {\begin{tabular}{llrrrr}
        \toprule
        \multirow{2}{*}{Environments} & \multirow{2}{*}{\makecell{Evaluation\\Conditions}} 
        & \multicolumn{1}{c}{\multirow{2}{*}{Offline}}
        & \multicolumn{3}{c}{$q_\text{ada}$} \\
        \cmidrule(lr){4-6}
        &&& \multicolumn{1}{c}{$1\text{M}$} 
        & \multicolumn{1}{c}{$2\text{M}$} & \multicolumn{1}{c}{$3\text{M}$} \\
        \midrule
        \multirow{3}{*}{Hopper-v2}
        &Normal     &$107.6\pm0.7$&$ 81.0\pm1.2$&$ 75.7\pm1.3$&$ 73.6\pm1.2$ \\ 
        &Random     &$ 46.9\pm1.8$&$ 64.3\pm1.2$&$ 68.9\pm1.4$&$ 66.1\pm1.3$ \\ 
        &Adversarial&$ 15.8\pm0.1$&$ 43.0\pm0.8$&$ 54.5\pm1.3$&$ 66.9\pm1.2$ \\ 
        \midrule
        \multirow{3}{*}{HalfCheetah-v2}
        &Normal     &$ 86.8\pm0.7$&$ 89.7\pm0.2$&$ 90.6\pm0.3$&$ 93.5\pm0.4$ \\ 
        &Random     &$ 61.2\pm1.1$&$ 74.5\pm0.5$&$ 73.6\pm0.5$&$ 75.5\pm0.6$ \\ 
        &Adversarial&$  4.9\pm0.1$&$ 63.0\pm0.3$&$ 61.7\pm0.4$&$ 65.9\pm0.5$ \\
        \midrule
        \multirow{3}{*}{Ant-v2}
        &Normal     &$113.2\pm1.7$&$117.2\pm1.4$&$129.1\pm1.1$&$129.9\pm1.1$ \\ 
        &Random     &$ 82.9\pm1.8$&$ 95.0\pm1.4$&$ 96.5\pm1.6$&$100.2\pm1.8$ \\ 
        &Adversarial&$-24.0\pm1.2$&$ 97.2\pm0.7$&$112.2\pm0.5$&$113.9\pm1.0$ \\
        \bottomrule
    \end{tabular}}
    \label{tab:results_q_adaptive_iql}
\end{table}

\section{Comparison of Fine-Tuning under Different Perturbation Generation Mechanisms}
The main text studies adversarial fine-tuning using a static perturbation set generated in advance by Differential Evolution (DE).
This setting employs the adaptive perturbation probability $q_\text{ada}$.
In this appendix, we examine how the adaptation behavior changes when the perturbation generation mechanism differs.
To this end, we introduce step-wise adversarial fine-tuning using Projected Gradient Descent (PGD)~\cite{pgd}, where perturbations are generated at each interaction step.

Algorithm~\ref{alg:pgd_attack} summarizes the PGD procedure.
In the PGD setting, at each environment step, a perturbation $\vect{\delta}$ is iteratively updated to reduce the critic value $Q(\vect{s},\vect{a})$ for the current state $\vect{s}$ and action $\vect{a}$.
We use $K=5$ update steps and a step size of $\alpha=0.1$.
The perturbation bound is set to $\epsilon=0.3$ for Hopper-v2 and HalfCheetah-v2, and $\epsilon=0.5$ for Ant-v2.
We use the same values during evaluation.
Fine-tuning runs for 3M environment steps.
All other hyperparameters remain identical to those in the main experiments.
DE generates a static perturbation set before fine-tuning.
During training, the same perturbation instance applies throughout each episode.
In contrast, PGD generates perturbations at every step based on critic gradients.
Both methods share the same perturbation magnitude constraint.
Therefore, the comparison primarily aims to isolate the effect of the perturbation generation mechanism rather than perturbation strength.
We evaluate policies under three conditions: Normal (no perturbation), Adversarial (static perturbation set), and PGD-Adv (step-wise PGD attack during evaluation).
Under PGD-Adv, perturbations are generated by applying PGD to the pretrained model.
This design aligns the evaluation condition with the DE setting, where perturbations are also derived from the pretrained model.

Table~\ref{tab:results_qada_onlineadv} reports the results.
Across all environments, fine-tuned models substantially improve performance under adversarial conditions compared to the pretrained models.
However, the improvement pattern depends on the perturbation generation mechanism used during fine-tuning.
Models fine-tuned with static perturbations tend to achieve higher performance under static adversarial evaluation.
Models fine-tuned with PGD tend to achieve higher performance under PGD-Adv evaluation.
This trend appears consistently across all environments.
Differences also emerge in the compatibility with performance under the Normal condition.
Static perturbation fine-tuning, combined with adaptive $q$, tends to preserve performance under the Normal condition while improving adversarial robustness.
In contrast, PGD generates perturbations that minimize the critic value at each step.
As a result, policy updates emphasize stability against step-wise worst-case perturbations.
Such updates tend to shift action outputs toward more conservative behavior.
Robustness under perturbations increases, but slight deviations from the nominal optimal action may occur under the Normal condition.

These results suggest that robustness achieved by adversarial fine-tuning depends not only on perturbation magnitude but also on the perturbation generation mechanism and its temporal structure.
The findings indicate that designing the perturbation model according to the expected disturbance characteristics in deployment environments plays a critical role.
Selecting a perturbation mechanism consistent with the evaluation condition may help achieve a better balance between robustness and nominal performance.
\label{appdx:pgd_adv}
\begin{algorithm}[t]
    \caption{Step-wise Adversarial Perturbation via Projected Gradient Descent (PGD)}
    \label{alg:pgd_attack}
    \begin{algorithmic}[1]
    \Require state $\vect{s}$, action $\vect{a}$, critic $Q(\cdot)$, step size $\alpha$, bound $\epsilon$, iterations $K$
    \Require initial perturbation $\vect{\delta}^{(0)}$
    
    \State $\vect{\delta} \gets \vect{\delta}^{(0)}$
    \For{$k = 1$ to $K$}
        \State Define perturbed action $\tilde{\vect{a}}(\vect{\delta}) \gets \vect{a} \odot ( \vect{1} + \vect{\delta})$
        \State Compute ascent direction on the perturbation:
        \[
            \vect{g} \gets \nabla_{\vect{\delta}} \, \mathcal{L}\!\left(\vect{s}, \tilde{\vect{a}}(\vect{\delta}); Q \right)
        \]
        \State Update perturbation toward the worst-case direction:
        \[
            \vect{\delta} \gets \vect{\delta} + \alpha \cdot \mathrm{Normalize}(\vect{g})
        \]
        \State Project onto the feasible set $\Delta \triangleq \{\vect{\delta} \mid \|\vect{\delta}\|_\infty \le \epsilon\}$:
        \[
            \vect{\delta} \gets \Pi_{\Delta}(\vect{\delta})
        \]
    \EndFor
    \State \Return $\vect{\delta}$
    \end{algorithmic}
\end{algorithm}
\begin{table}[H]
    \caption{
    Performance comparison between static DE-based adversarial fine-tuning with adaptive $q$ and step-wise PGD-based fine-tuning.
    \textit{Offline} denotes the pretrained policy without online fine-tuning.
    Columns $1\text{M}$, $2\text{M}$, and $3\text{M}$ report performance after the corresponding numbers of fine-tuning steps.
    Evaluation conditions include Normal, Adversarial (static perturbation set generated from the pretrained model), and PGD-Adv (step-wise PGD perturbations generated against the pretrained model).
    Values indicate D4RL-normalized episodic reward.
    }
    \centering
    {\begin{tabular}{llrrrrrrr}
        \toprule
        \multirow{2}{*}{Environments} & \multirow{2}{*}{\makecell{Evaluation\\Conditions}} 
        & \multicolumn{1}{c}{\multirow{2}{*}{Offline}} 
        & \multicolumn{3}{c}{$q_\text{ada}$} 
        & \multicolumn{3}{c}{PGD-adversarial} \\
        \cmidrule(lr){4-6}
        \cmidrule(lr){7-9}
        & & &\multicolumn{1}{c}{$1\text{M}$} 
        & \multicolumn{1}{c}{$2\text{M}$} & \multicolumn{1}{c}{$3\text{M}$} 
        & \multicolumn{1}{c}{$1\text{M}$} 
        & \multicolumn{1}{c}{$2\text{M}$} & \multicolumn{1}{c}{$3\text{M}$} \\
        \midrule
        \multirow{3}{*}{Hopper-v2}
        &Normal     &$111.1$&$100.2$&$ 75.0$&$ 97.5$&$ 93.0$&$100.5$&$ 79.2$ \\ 
        &Adversarial&$ 13.7$&$ 87.9$&$ 88.2$&$ 81.5$&$ 80.6$&$ 77.6$&$ 69.6$ \\ 
        &PGD-Adv &$ 11.8$&$ 67.1$&$ 58.1$&$ 73.0$&$ 81.7$&$ 90.5$&$ 85.1$ \\
        \midrule
        \multirow{3}{*}{HalfCheetah-v2}
        &Normal     &$ 96.6$&$ 89.0$&$ 93.8$&$ 93.0$&$ 68.9$&$ 69.3$&$ 70.1$ \\ 
        &Adversarial&$ 12.1$&$ 63.4$&$ 69.2$&$ 75.1$&$ 60.0$&$ 63.6$&$ 64.2$ \\
        &PGD-Adv &$ 35.2$&$ 50.0$&$ 51.7$&$ 51.1$&$ 56.1$&$ 61.5$&$ 61.7$\\
        \midrule
        \multirow{3}{*}{Ant-v2}
        &Normal     &$119.9$&$124.8$&$131.0$&$130.8$&$128.9$&$133.6$&$130.4$ \\ 
        &Adversarial&$-21.0$&$ 98.2$&$115.7$&$118.5$&$ 88.3$&$102.2$&$103.1$ \\
        &PGD-Adv &$ 72.1$&$ 45.2$&$ 53.8$&$ 49.6$&$ 91.4$&$101.3$&$ 97.1$\\
        \bottomrule
    \end{tabular}}
    \label{tab:results_qada_onlineadv}
\end{table}
\end{document}